\documentclass[a4paper,review,fleqn]{cas-dc}

\usepackage[numbers]{natbib}

\usepackage{listings}
\usepackage{booktabs}
\usepackage{pifont}
\newcommand{\cmark}{\ding{51}}
\newcommand{\xmark}{\ding{55}}
\usepackage{xurl}
\usepackage{multirow}

\usepackage{lineno}
\def\tsc#1{\csdef{#1}{\textsc{\lowercase{#1}}\xspace}}
\tsc{WGM}
\tsc{QE}
\tsc{EP}
\tsc{PMS}
\tsc{BEC}
\tsc{DE}

\begin{document}
\let\WriteBookmarks\relax
\def\floatpagepagefraction{1}
\def\textpagefraction{.001}
\shorttitle{Ontology-Driven Cyberbullying Behavior and Network Analysis}
\shortauthors{M.N. Hoque et~al.}

\title [mode = title]{From Perpetrators' Digital Footprints to Behavioral Communities: An Ontology-Driven Framework for Bengali Cyberbullying Analysis--- FULL VERSION}                      

\tnotetext[1]{This work was partially supported by the Higher Education Acceleration and Transformation (HEAT) project, funded by the World Bank and the Government of Bangladesh, through the sub-project no. PIN-13211 and the ICT Division, Bangladesh (Fellowship code number: 1280101-120008431-3821117).}


\author[1,2]{Md. Nesarul Hoque}

\credit{Writing – review \& editing, Writing – original draft, Visualization, Validation, Software, Methodology, Investigation, Formal analysis, Data curation, Conceptualization}


\author[1]{Md Hanif Seddiqui}

\credit{Writing – review \& editing, Supervision, Resources, Methodology, Conceptualization}

\author[1]{Miskatul Anwar}

\credit{Writing – review \& editing, Visualization, Validation, Software}

\author[1]{Rudra Pratap Deb Nath}
[orcid=0000-0001-8807-1927]
\cormark[1]
\ead{rudra@cu.ac.bd}

\credit{Writing – review \& editing, Writing – original draft, Supervision, Resources, Project administration, Methodology, Formal analysis, Investigation, Funding acquisition, Conceptualization}

\affiliation[1]{organization={Big Data, Information and Knowledge Engineering Lab, Department of Computer Science and Engineering, University of Chittagong},
                city={Chattogram},
                postcode={4331}, 
                country={Bangladesh}}

\affiliation[2]{organization={Department of Computer Science and Engineering, Gopalganj Science and Technology University},
                city={Gopalganj},
                postcode={8105}, 
                country={Bangladesh}}


\cortext[cor1]{Corresponding author}


\begin{abstract}
Cyberbullying involves behavioral and contextual patterns beyond isolated abusive messages. However, existing approaches remain fragmented across content detection, semantic representation, and network analysis, with limited interpretable perpetrator representations, multi-scale structural modeling, and explanations of behavior-driven similarity and influence propagation. These limitations hinder the identification of persistent perpetrators, characterization of behavioral communities, and assessment of potentially influential users. This paper proposes Ontology-Driven Cyberbullying Behavior and Network Analysis (ODCBNA), a perpetrator-centric framework integrating requirements-driven ontology design, knowledge graph construction, behavioral modeling, community analysis, and influence modeling. Using the Bengali Contextual and Diverse Cyberbullying (BenCoDiC) corpus as a use case, ODCBNA identifies $1{,}063$ users with at least three bullying comments or replies as perpetrators and represents each through a $64$-dimensional behavioral and contextual feature vector. Cosine similarity and $k$-nearest-neighbor modeling produce a weighted network with $27{,}554$ edges. Louvain analysis identifies six communities with a modularity of approximately $0.495$ and $27$ sub-communities. Feature-contribution analysis identifies the Video, Impact, and User Profile groups as the strongest contributors to behavioral similarity and community separation, while contextual dimensions support interpretation. Community-constrained influence analysis shows that IM Greedy achieves small positive gains in all six communities, although different seed-selection methods produce comparable diffusion spreads despite limited seed overlap. These findings provide an interpretable basis for understanding perpetrator behavior, community organization, and potential influence patterns in Bengali online environments.
\end{abstract}




\maketitle

\section{Introduction}
\label{sec:intro}

Cyberbullying is a critical concern on online social platforms because harmful online communication can lead to serious psychological and social consequences. Prior studies have associated cyberbullying victimization with depression, anxiety, suicidal ideation, and related adverse outcomes~\cite{Kowalski2014Bullying, Hinduja2010Bullying}. A widely documented public example is the online harassment directed at Olympic boxer Imane Khelif during the Paris 2024 Olympics, where false claims and identity-based attacks spread across social media and led to a legal complaint for online harassment~\cite{Khelif2024AP}. Such incidents illustrate that cyberbullying may appear through individual posts or comments, but the underlying behavior often involves repeated participation, recurring targets, and interaction with other users. Its meaning and severity may also depend on language, discussion context, topic, and social environment. Therefore, understanding cyberbullying requires attention not only to abusive content but also to the behavioral and contextual patterns surrounding abusive interactions~\cite{Hoque2023Challenges, Ribeiro2018Characterizing}.

This need becomes especially clear in politically sensitive Bengali discussions on YouTube. A classifier may identify an abusive comment under a news video, but that decision says little about whether the same user repeatedly posts harmful content, targets similar individuals or groups, reacts to similar political contexts, or belongs to a cluster of users with comparable behavior. The comment is therefore only the visible trace of a wider digital footprint. A content-level approach can detect the harmful message, but it cannot explain how abusive users are behaviorally connected, how their activity varies across contexts, or which users may become structurally important within a community. This motivates a perpetrator-centric analysis that connects abusive content with user behavior, contextual exposure, community organization, and influence patterns.

Recent studies and surveys show that content-based cyberbullying detection has expanded across machine learning, deep learning, transformer-based models, feature engineering, multilingual datasets, and platform-specific settings~\cite{Mali2025Automatic, Hoque2024Exploring, Mishra2024Shielding, Philipo2026Cyberbullying}. Despite these advances, most content-centered approaches remain focused on determining whether individual posts, comments, or messages contain abusive content. They therefore provide limited insight into persistent perpetrators, multiple bullying-related interactions, contextual exposure, or the position of abusive users within broader behavioral structures. Contextual ambiguity and variations in user behavior further limit approaches that rely primarily on message content~\cite{Fortuna2018Survey, Hoque2023Challenges}.

Alongside content-centered detection, ontology-driven and semantic representation methods organize heterogeneous social-media entities, attributes, and relationships within explicit domain models~\cite{Grigaliunas2023Ontology, Chun2024Ontology}. Such representations provide a structured way to connect digital artifacts, user-related attributes, contextual information, and cyber-violence concepts within a consistent semantic framework. However, semantic representation alone does not reveal how abusive users are behaviorally similar, how they organize into communities, or how influence may propagate within those communities.

Graph- and network-aware studies address the relational side of cyberbullying and harmful online behavior. User-centric analysis shows that hateful or abusive users may differ in activity patterns, vocabulary, and network structure~\cite{Ribeiro2018Characterizing}. Work on online trolling further shows that discussion context and exposure to prior harmful posts can increase subsequent antisocial behavior~\cite{Cheng2017Troll}. Recent cyberbullying studies also explore fine-grained role identification and GNN-based detection by incorporating behavioral features or graph-structured relations among text instances~\cite{Wang2025Role, Maity2024MTBullyGNN}. These studies demonstrate the value of relational and behavioral signals, but they remain mainly oriented toward detection, role identification, or representation learning.

Overall, existing research remains divided among content analysis, semantic representation, and network modeling. As a result, cyberbullying is rarely examined through a single framework that connects meaning, perpetrator behavior, structural organization, and influence processes. This separation makes it difficult to explain how abusive users can be represented, compared, grouped, and examined for their potential influence. We identify four research gaps (RGs):

\begin{itemize}

\item[RG1] \textbf{Fragmented modeling paradigms:}
Content classification, graph analysis, and semantic representation are usually studied separately~\cite{Fortuna2018Survey, Ribeiro2018Characterizing}. The lack of integration limits multi-level analysis of cyberbullying.

\item[RG2] \textbf{Lack of interpretable behavioral abstraction:}
Most methods rely on textual features or basic user statistics rather than interpretable, multi-dimensional representations of perpetrators. This limits the analysis of abusive activity across multiple interactions~\cite{Hoque2023Challenges}.

\item[RG3] \textbf{Limited multi-scale structural modeling:}
Previous studies examine user networks and interaction patterns, but similarity-based network construction and hierarchical community organization remain insufficiently explored~\cite{Ribeiro2018Characterizing}. Consequently, the structure of cyberbullying behavior across different scales is not well understood.

\item[RG4] \textbf{Lack of interpretable behavior-driven similarity and diffusion analysis:}
Network-based studies indicate that abusive behavior may spread through interactions~\cite{Cheng2017Troll}, but they rarely clarify how behavioral attributes contribute to user similarity or how different users influence propagation. This limits the interpretability of network findings and their use in targeted intervention.

\end{itemize}

To address these gaps, we propose Ontology-Driven Cyberbullying Behavior and Network Analysis (ODCBNA), a perpetrator-centric framework for YouTube cyberbullying analysis. Guided by analytical requirements and competency questions, ODCBNA defines a cyberbullying application ontology as the semantic foundation for knowledge graph construction, behavioral modeling, community analysis, and influence modeling. The framework is instantiated using the Bengali Contextual and Diverse Cyberbullying (BenCoDiC) corpus, which provides labeled YouTube political-news interactions and associated behavioral and contextual metadata~\cite{Hoque2026BenCoDiC}. By integrating these analytical components, ODCBNA supports interpretable investigation of perpetrator behavior, community organization, and potential influence. The following research contributions (RCs) correspond directly to RG1--RG4, respectively:

\begin{itemize}

\item[RC1] \textbf{Ontology-driven analytical framework:}
We develop an integrated pipeline connecting competency-question-guided ontology design, knowledge graph construction, heterogeneous graph transformation, and network analysis. This integration preserves semantic relationships while supporting behavioral, contextual, and structural analysis within a unified framework.

\item[RC2] \textbf{Perpetrator-centric behavioral--contextual representation:}
We construct an interpretable 64-dimensional representation covering User Profile, Video, Impact, Language, Target, Party, Time, and Channel. The representation enables consistent comparison among perpetrators and supports analysis of the behavioral and contextual dimensions associated with community structure.

\item[RC3] \textbf{Multi-scale network modeling:}
We construct a weighted behavioral-similarity network and identify six communities and $27$ sub-communities. This hierarchical structure reveals perpetrator organization at multiple scales and supports finer analysis within the detected communities.

\item[RC4] \textbf{Interpretable similarity and influence analysis:}
We develop a behavioral contribution framework and a community-constrained influence maximization procedure. The results show that the Video, Impact, and User Profile groups provide the strongest contributions to similarity, while largely distinct seed sets achieve closely comparable diffusion outcomes under the adopted IC settings.

\end{itemize}

Together, these contributions connect semantic representation, perpetrator behavior, community structure, and influence analysis, providing a broader account of cyberbullying beyond isolated content.

The rest of this paper is organized as follows. Section~\ref{sec:related-work} reviews related studies, Section~\ref{sec:methodology} describes the proposed framework, and Section~\ref{sec:experimental-evaluation} reports the experimental analysis and findings. Section~\ref{sec:conclusion} concludes the paper.

\section{Related Work}
\label{sec:related-work}

Cyberbullying research broadly follows three directions: content-based detection, ontology-driven semantic modeling, and graph- or network-based analysis. Each direction contributes useful methods, but they are often examined separately. This separation limits efforts to connect the semantic, behavioral, and structural aspects of cyberbullying. This section reviews representative studies from these areas and clarifies the position of the proposed framework.

\subsection{Cyberbullying Detection}
\label{subsec:cb-detect-approach}

Much of the existing research treats cyberbullying detection as a supervised text classification task. Recent surveys show that machine learning, deep learning, and transformer models are widely used and perform well on benchmark datasets~\cite{Philipo2026Cyberbullying, Mishra2024Shielding}. Hybrid architectures that combine CNNs, recurrent networks, transformer representations, attention, ensemble learning, and adversarial training have also improved accuracy, robustness, and cross-domain performance~\cite{Harshitha2024ProTect, Cuzzocrea2025Cyberbullying, Xiong2025Topic}.

Research has also expanded to low-resource and multilingual settings. Atif et al.~\cite{Atif2024Cyberbullying} use user-level signals for Roman Urdu cyberbullying detection, whereas Hoque et al.~\cite{Hoque2026Advancing} develop a transformer-stacking framework for Bengali. Dataset-oriented studies have adopted broader definitions of cyberbullying by considering repetition, peerness, and intent to harm~\cite{Ejaz2024Towards}.

Even with this progress, most methods remain centered on content detection or user classification. They offer limited evidence about how abusive users form behavioral communities, how their interaction patterns develop, or how cyberbullying operates as a networked social process.

\subsection{Ontology-Driven Semantic Approaches}
\label{subsec:onto-approach}

Another line of research uses ontology-driven modeling to organize heterogeneous social media data and represent domain knowledge. Ontologies define entities, relationships, and contextual information in a structured form, allowing information from different sources to be linked semantically~\cite{DebNathAspect2020}.

Grigaliūnas et al.~\cite{Grigaliunas2023Ontology} propose an ontology-driven digital profiling framework for linking social media evidence, while Chun et al.~\cite{Chun2024Ontology} develop a comprehensive ontology for cyber violence victimization, capturing participants, behavioral characteristics, and outcomes.

Although these approaches support semantic representation and structured profiling, their integration with perpetrator-level behavioral similarity networks, hierarchical community analysis, and influence propagation remains limited.

\subsection{Graph- and Network-Aware Cyberbullying Analysis}
\label{subsec:graph-network-approach}

To address the relational nature of cyberbullying, recent work incorporates graph-based and network-aware modeling. Surveys highlight increasing use of interaction networks, contextual signals, and relational structures for cyberbullying analysis~\cite{Mishra2024Shielding, Philipo2026Cyberbullying}.

Some studies move beyond content classification toward behavioral modeling. Wang et al.~\cite{Wang2025Role} propose a multi-level role identification framework that captures fine-grained user roles and their evolution. Other approaches explore distributed and privacy-preserving cyberbullying detection through transfer learning and federated learning frameworks~\cite{Kumari2025Towards}. More recently, graph neural network (GNN)-based approaches have been explored to capture interaction structures and contextual dependencies in cyberbullying data~\cite{Maity2024MTBullyGNN}.

Although these approaches incorporate contextual, behavioral, or relational information, they primarily emphasize detection or role identification. Their network representations do not systematically combine perpetrator-level behavioral similarity, hierarchical community analysis, and influence propagation within a unified analytical framework.

\textbf{Summary and Discussion.}
Existing studies have advanced cyberbullying detection, semantic representation, behavioral profiling, and relational modeling. However, these capabilities are generally addressed separately, with limited integration of interpretable perpetrator representations, multi-scale behavioral networks, and influence propagation analysis. Consequently, a unified approach connecting semantic knowledge, perpetrator behavior, structural organization, and diffusion remains insufficiently explored.

Table~\ref{tab:related-comparison} compares representative studies with ODCBNA using semantic, behavioral, structural, and analytical criteria. Behavioral representation corresponds to RG2, similarity and community modeling to RG3, and explainability and influence modeling to RG4. RG1 is reflected in the integration of these capabilities rather than a separate criterion. ODCBNA integrates competency-question-guided ontology design, knowledge graph construction, perpetrator-level behavioral modeling, hierarchical community analysis, and interpretable similarity and influence modeling within a unified framework.

\begin{table*}[ht]
\centering
\caption{Comparative analysis of existing studies and ODCBNA across semantic, behavioral, structural, and analytical dimensions.}
\label{tab:related-comparison}
\footnotesize
\setlength{\tabcolsep}{4pt}
\begin{tabular}{lccccc cc}
\toprule
\multirow{2}{*}{\textbf{Ref.}}
& \multicolumn{2}{c}{\textbf{Semantic / Structural Basis}}
& \textbf{RG2}
& \multicolumn{2}{c}{\textbf{RG3}}
& \multicolumn{2}{c}{\textbf{RG4}} \\
\cmidrule(lr){2-3} \cmidrule(lr){5-6} \cmidrule(lr){7-8}
& \textbf{Ontology} & \textbf{Perp-Graph}
& \textbf{Behav. Rep.}
& \textbf{Similarity} & \textbf{Community}
& \textbf{Explainability} & \textbf{Influence} \\
\midrule
Ejaz et al.~\cite{Ejaz2024Towards}
& \xmark & \xmark & \xmark & \xmark & \xmark & \xmark & \xmark \\

Harshitha et al.~\cite{Harshitha2024ProTect}
& \xmark & \xmark & \xmark & \xmark & \xmark & \xmark & \xmark \\

Cuzzocrea et al.~\cite{Cuzzocrea2025Cyberbullying}
& \xmark & \xmark & \xmark & \xmark & \xmark & \xmark & \xmark \\

Xiong et al.~\cite{Xiong2025Topic}
& \xmark & \xmark & \xmark & \xmark & \xmark & \cmark & \xmark \\

Atif et al.~\cite{Atif2024Cyberbullying}
& \xmark & \xmark & \cmark & \xmark & \xmark & \xmark & \xmark \\

Hoque et al.~\cite{Hoque2026Advancing}
& \xmark & \xmark & \xmark & \xmark & \xmark & \xmark & \xmark \\

Wang et al.~\cite{Wang2025Role}
& \xmark & \xmark & \cmark & \xmark & \xmark & \xmark & \xmark \\

Kumari and Kaur~\cite{Kumari2025Towards}
& \xmark & \xmark & \xmark & \xmark & \xmark & \cmark & \xmark \\

Maity et al.~\cite{Maity2024MTBullyGNN}
& \xmark & \xmark & \xmark & \cmark & \xmark & \xmark & \xmark \\

Grigaliūnas et al.~\cite{Grigaliunas2023Ontology}
& \cmark & \xmark & \cmark & \cmark & \xmark & \cmark & \xmark \\

Chun et al.~\cite{Chun2024Ontology}
& \cmark & \xmark & \xmark & \xmark & \xmark & \cmark & \xmark \\

\midrule
\textbf{Proposed ODCBNA framework}
& \cmark & \cmark & \cmark & \cmark & \cmark & \cmark & \cmark \\
\bottomrule
\end{tabular}

\vspace{2pt}
\parbox{\textwidth}{\footnotesize
\textbf{Notes:}
Perp-Graph = Perpetrator-level graph;
Behav. Rep. = User-level behavioral representation.
Similarity includes explicit text- or user-level similarity modeling.
Explainability includes model explanations, feature attribution, or interpretable semantic profiling.
Influence refers to network diffusion or influence modeling.
\cmark\ indicates a reported capability, while
\xmark\ indicates that the capability is not a principal component of the study.
}
\end{table*}

\section{Methodology}
\label{sec:methodology}

\begin{figure*}[ht]
\centering
\includegraphics[width=\textwidth]{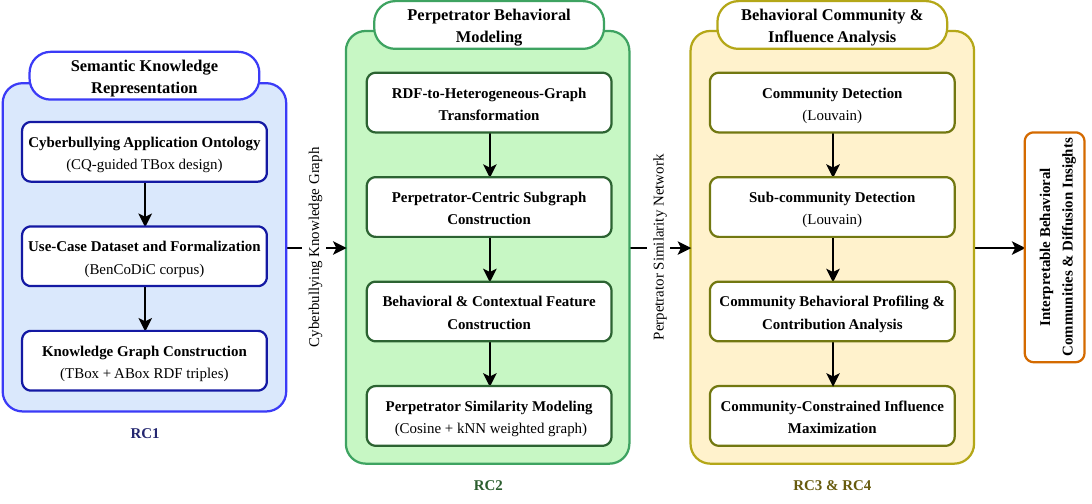}
\caption{Overview of the ODCBNA framework.}
\label{fig:overall-pipeline}
\end{figure*}

This section presents the Ontology-Driven Cyberbullying Behavior and Network Analysis (ODCBNA) framework for analyzing cyberbullying from a perpetrator-centric perspective. The problem is to identify users exhibiting multiple bullying-labeled actions, characterize their observed behavioral patterns, examine their community organization, and model potential influence propagation. Guided by these analytical requirements and the corresponding competency questions (CQs), ODCBNA begins with the design of a cyberbullying application ontology that provides the semantic foundation for knowledge graph construction, behavioral modeling, community analysis, and influence modeling. Fig.~\ref{fig:overall-pipeline} illustrates the overall workflow of ODCBNA. The framework is instantiated and evaluated using the Bengali Contextual and Diverse Cyberbullying (BenCoDiC) corpus as a use case~\cite{Hoque2026BenCoDiC}.

\subsection{Semantic Knowledge Representation}
\label{subsec:semantic-representation}

The Semantic Knowledge Representation stage establishes the semantic foundation for perpetrator-centric analysis through three components: Cyberbullying Application Ontology, Use-Case Dataset and Formalization, and Knowledge Graph Construction. First, a competency-question-guided application ontology defines the concepts, properties, and relationships required for the analysis. Next, the BenCoDiC corpus is introduced as a use case to formalize observed interactions and their associated behavioral and contextual information. Finally, the ontology is instantiated using the corpus records to construct a cyberbullying knowledge graph for subsequent behavioral modeling and network analysis.

\subsubsection{Cyberbullying Application Ontology}
\label{subsubsec:cb-onto}

Perpetrator-centric analysis of YouTube cyberbullying requires a structured representation of users, channel content, interactions, and their associated behavioral and contextual information. If these entities and attributes are processed as independent records or isolated features, their semantic relationships may be lost. For example, a bullying comment is not only a text instance. It is authored by a user, associated with channel content, expressed in a language, published at a time, and may target a particular identity or receive engagement responses. Therefore, an explicit semantic model is required to represent these entities and their relationships in a consistent and machine-processable form.

To meet this need, we define a domain-specific application ontology guided by the analytical requirements of ODCBNA. The ontology serves as a semantic schema for representing entities, attributes, and relationships relevant to perpetrator-centric cyberbullying analysis. Its explicit definitions preserve semantic relationships among users, interactions, textual content, channel content, and contextual information. These relationships can be represented through RDF triples, providing a consistent foundation for knowledge graph construction and subsequent behavioral and network analyses.

The analytical requirements are expressed through the following competency questions, which define the intended scope of the framework:

\begin{itemize}
\item[\textbf{CQ1}] Which users exhibit multiple bullying-labeled actions across interactions?
\item[\textbf{CQ2}] What behavioral characteristics define different types of perpetrators?
\item[\textbf{CQ3}] What structural patterns characterize the perpetrator similarity network at multiple scales?
\item[\textbf{CQ4}] How are perpetrators structurally organized into communities and sub-communities based on behavioral similarity?
\item[\textbf{CQ5}] How are behavioral patterns associated with contextual environments across communities?
\item[\textbf{CQ6}] Which behavioral dimensions contribute most to similarity and community formation?
\item[\textbf{CQ7}] Which users are most influential within each community under a diffusion process?
\item[\textbf{CQ8}] How does influence propagate across structurally similar perpetrators?
\end{itemize}

These CQs guide the selection of ontology classes and properties by identifying the information required for perpetrator-centric analysis. In particular, the ontology must represent user behavior, textual labels, interaction context, content exposure, target identity, discussion subjects, temporal information, and engagement impact. While the ontology establishes the semantic foundation, the subsequent behavioral and network analysis components address the computational aspects of these questions.

\begin{figure*}[ht]
\centering
\includegraphics[width=\textwidth]{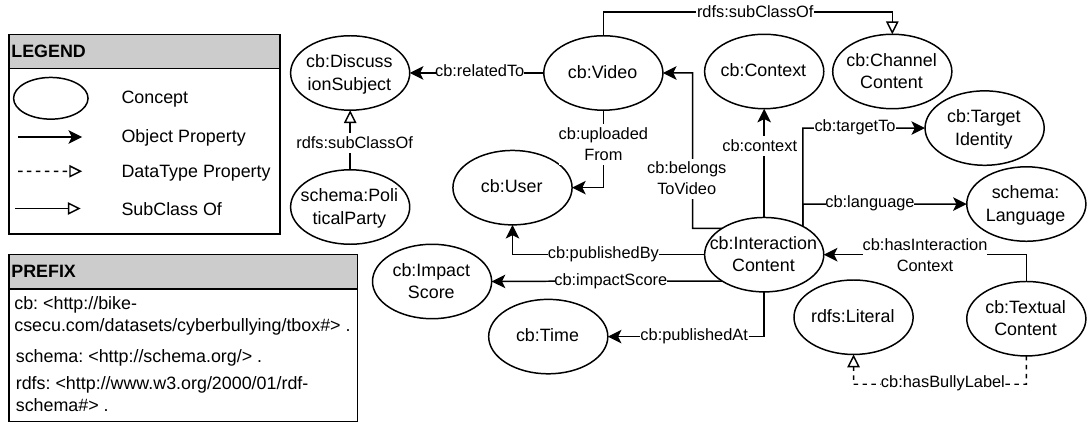}
\caption{High-level structure of the cyberbullying application ontology (TBox), including subclass relationships and selected datatype properties.}
\label{fig:cb-ontology}
\end{figure*}

Fig.~\ref{fig:cb-ontology} shows the high-level structure of the proposed cyberbullying application ontology. The ontology combines a domain-specific vocabulary in the \texttt{cb:} namespace with selected terms from established vocabularies, including Schema.org (\texttt{schema:}) and RDF Schema (\texttt{rdfs:}). In semantic knowledge representation, a knowledge base is commonly described using two components: the TBox and the ABox~\cite{Baader2007Description}. The TBox, or terminological component, defines the domain schema in terms of concepts, properties, and their semantic relationships, whereas the ABox, or assertional component, contains instance-level assertions that instantiate the TBox. Accordingly, the proposed ontology defines the TBox of the cyberbullying knowledge graph, while observed data provide the corresponding ABox through RDF triples.

The central concept in the ontology is
\texttt{cb:}\allowbreak\texttt{InteractionContent},
which represents an online interaction such as a comment or reply. The \texttt{cb:publishedBy} property links each interaction to its authoring \texttt{cb:User}, while \texttt{cb:belongsToVideo} links it to the relevant \texttt{cb:Video}. Video uploader accounts are connected through \texttt{cb:uploadedFrom}. To represent channel-published content more generally, \texttt{cb:ChannelContent} is defined as a parent class of \texttt{cb:Video} through \texttt{rdfs:subClassOf}. This hierarchy supports extensibility to other YouTube content types while preserving the video-specific relationships used in the present analysis. These relationships also support the distinction between abusive participants and content producers during perpetrator-centric subgraph construction.

The textual content associated with an interaction is represented by \texttt{cb:TextualContent}, with its cyberbullying annotation stored through the \texttt{cb:hasBullyLabel} datatype property. This separation keeps the label attached to the textual content while preserving its broader interaction context. The label indicates whether bullying is present, while the associated interaction identifies who posted the content, where it appeared, and its surrounding context.

The ontology also represents the contextual and behavioral dimensions required for downstream analysis. The \texttt{schema:Language} concept captures the linguistic form of an interaction, while \texttt{cb:TargetIdentity} represents the identity category targeted by the content. The \texttt{cb:DiscussionSubject} concept represents subjects associated with channel content through \texttt{cb:relatedTo}, with \texttt{schema:PoliticalParty} defined as its subclass to represent political discussion contexts. Temporal information is represented by \texttt{cb:Time} through \texttt{cb:publishedAt}, while audience response is represented by \texttt{cb:ImpactScore} through \texttt{cb:impactScore}. In addition, \texttt{cb:Context} captures interaction-level contextual information and is linked to \texttt{cb:InteractionContent} through \texttt{cb:context}.

Together, these concepts and relationships provide a semantic foundation for organizing cyberbullying-related information and supporting subsequent behavioral modeling and network analysis.

\subsubsection{Use-Case Dataset and Formalization}
\label{subsubsec:dataset}

To instantiate and evaluate the proposed ODCBNA framework, we employ the Bengali Contextual and Diverse Cyberbullying (BenCoDiC) corpus as a use case~\cite{Hoque2026BenCoDiC}. BenCoDiC is a context-rich Bengali cyberbullying resource developed from political news discourse on YouTube. The corpus is particularly suitable for perpetrator-centric analysis because it provides persistent user-level information together with multidimensional behavioral and contextual metadata covering users, videos, textual content, interaction context, time, engagement, political-party context, target identities, and language categories. It comprises $50{,}250$ textual instances associated with $100$ YouTube videos and $37{,}813$ user accounts. To protect user privacy, original user IDs, handlers, and channel titles are anonymized, while persistent anonymized identifiers preserve user-level continuity across interactions without exposing the original identities.

Let $\mathcal{I}$ denote the set of interaction instances, $\mathcal{U}$ the set of user accounts, $\mathcal{V}$ the set of videos, $\mathcal{TC}$ the set of textual content instances, $\mathcal{C}_{\mathrm{tx}}$ the set of contextual records, $\mathcal{T}$ the set of temporal records, $\mathcal{IS}$ the set of impact-score records, $\mathcal{PP}$ the set of political-party records, $\mathcal{L}$ the set of language records, and $\mathcal{TI}$ the set of target-identity records. In this use case, video and political-party records correspond to instances of \texttt{cb:Video} and \texttt{schema:PoliticalParty}, which are subclasses of \texttt{cb:ChannelContent} and \texttt{cb:DiscussionSubject}, respectively. Since each labeled textual instance is linked to its corresponding interaction and associated contextual records, the set of observed record combinations is represented as

\begin{equation*}
\mathcal{X}\subseteq
\mathcal{I}\times\mathcal{U}\times\mathcal{V}\times\mathcal{TC}
\times\mathcal{C}_{\mathrm{tx}}\times\mathcal{T}
\times\mathcal{IS}\times\mathcal{PP}\times\mathcal{L}\times\mathcal{TI}.
\label{eq:annotation-domain}
\end{equation*}

The binary label-assignment mapping is represented as

\[
f:\mathcal{X}\rightarrow \{\textit{bully}, \textit{notBully}\}.
\]

Here, $(i,u,v,tc,c,t,is,pp,l,ti)\in\mathcal{X}$ denotes the interaction, user, video, textual content, context, time, impact-score, political-party, language, and target-identity records, respectively. Accordingly, $f(i,u,v,tc,c,t,is,pp,l,ti)$ denotes the binary cyberbullying label associated with textual instance $tc$ and its linked records. The annotation decision is based on the textual content together with its available conversational context, while the remaining dimensions preserve the associated information required for subsequent behavioral analysis. For later use, $\ell(tc)$ denotes the label associated with textual instance $tc$ through $f$.

The final BenCoDiC labels were obtained through a two-tier human annotation process supported by sociolinguist-guided training. Three annotators independently labeled each sample at Level-1, followed by verification by three NLP domain experts at Level-2. The final annotations achieved Fleiss' $\kappa=0.86$ and Krippendorff's $\alpha=0.86$, indicating strong inter-annotator agreement.

These linked records and annotations provide the instance-level information required for ontology-guided knowledge graph construction, supporting subsequent perpetrator-centric behavioral representation, community analysis, and influence modeling.

\subsubsection{Knowledge Graph Construction}
\label{subsubsec:kg-construction}

Based on the application ontology introduced in Section~\ref{subsubsec:cb-onto}, the BenCoDiC corpus is transformed into a cyberbullying knowledge graph. The ontology provides the TBox, i.e., the schema-level concepts and properties, while the corpus provides the observed entities, relationships, and literal values required to construct the ABox. Accordingly, this step generates corpus-derived ABox triples by mapping BenCoDiC records and metadata fields to the corresponding classes and properties defined in the TBox, using Resource Description Framework (RDF) representation.

Let $\mathsf{I}$, $\mathsf{B}$, and $\mathsf{L}$ denote the sets of Internationalized Resource Identifiers, blank nodes, and literals, respectively. An RDF triple is represented as $(s,p,o)$, where
$s\in(\mathsf{I}\cup\mathsf{B})$ is a subject resource,
$p\in\mathsf{I}$ is a predicate, and
$o\in(\mathsf{I}\cup\mathsf{B}\cup\mathsf{L})$ is an object resource, blank node, or literal value~\cite{Cyganiak2014RDF}. Accordingly, the cyberbullying knowledge graph is represented as a finite set of RDF triples:

\[
\mathcal{G}_{KG}\subseteq
(\mathsf{I}\cup\mathsf{B})
\times
\mathsf{I}
\times
(\mathsf{I}\cup\mathsf{B}\cup\mathsf{L}).
\]

Within $\mathcal{G}_{KG}$, the schema-level triples correspond to the TBox, while the corpus-derived instance-level triples constitute the ABox.

\begin{figure}
\centering
\includegraphics[width=1.0\columnwidth]{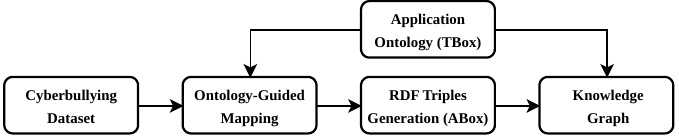}
\caption{Ontology-driven knowledge graph construction pipeline.}
\label{fig:kg-construction}
\end{figure}

Fig.~\ref{fig:kg-construction} illustrates the ontology-driven knowledge graph construction pipeline. The process starts with the cyberbullying dataset and the application ontology. The ontology provides the target semantic schema, while the dataset provides the observed entities, relationships, and literal values. Through ontology-guided mapping, corpus records are converted into RDF triples that instantiate the TBox~\cite{DebNath2021, DebNath2017}.

In this mapping process, dataset entities such as users, videos, interaction content instances, textual content instances, languages, target identities, political contexts, temporal records, and impact scores are assigned resource identifiers and instantiated according to the corresponding TBox classes. Relationships among these entities are encoded using object properties defined in the TBox. For example, an interaction is linked to the user who published it, the video under which it appears, its textual content, its language, its target identity, its publication time, and its impact-related information. Literal values such as textual content, timestamps, bullying labels, and engagement statistics are encoded using datatype properties.

The generated ABox triples, together with the RDF representation of the TBox, form the cyberbullying knowledge graph. This graph provides a unified multi-relational representation of cyberbullying interactions by explicitly linking abusive textual content with users, videos, contextual metadata, and engagement signals. The resulting knowledge graph serves as the semantic foundation for the subsequent RDF-to-heterogeneous-graph transformation and perpetrator-centric behavioral modeling.

\subsection{Perpetrator Behavioral Modeling}
\label{subsec:behavioral-modeling}

This stage transforms the ontology-grounded knowledge graph into a perpetrator-centric behavioral representation suitable for network analysis. It includes heterogeneous-graph transformation, perpetrator-centric subgraph construction, behavioral and contextual feature construction, and similarity modeling to produce the weighted perpetrator similarity network.

\subsubsection{RDF-to-Heterogeneous-Graph Transformation}
\label{subsubsec:rdf-hetero-graph-transform}

While RDF provides a flexible semantic representation, downstream graph learning and network analysis require a computationally efficient heterogeneous graph with explicit typing and indexing~\cite{Shi2017survey}. Therefore, the RDF knowledge graph is transformed into a heterogeneous graph through an intermediate tabular representation consisting of typed node tables, typed edge tables, and type-specific identifier mappings.

Let $G_h=(\mathcal{N}_h,\mathcal{E}_h)$ denote the resulting heterogeneous graph, where $\mathcal{N}_h$ is the set of typed nodes and $\mathcal{E}_h$ is the set of typed edges. Each typed edge is represented as $(n_s,n_t,r)\in\mathcal{E}_h$, where $n_s\in\mathcal{N}_h$ is the source node, $n_t\in\mathcal{N}_h$ is the target node, and $r$ denotes the relation type derived from the ontology.

The transformation proceeds in three stages. First, RDF triples are loaded from the Turtle files, and node instances are identified based on ontology classes and URI patterns. For each node type, relevant literal attributes and relational references are extracted and organized into node-specific tables.

Second, type-specific identifier mappings are constructed to convert URI-based entities into contiguous integer indices, enabling efficient storage and computation in graph-based frameworks.

Third, typed edge tables are generated by extracting relations defined in the ontology between compatible node types. These tables encode multi-relational connections while preserving the semantic structure of the original RDF graph.

\begin{figure}
\centering
\includegraphics[width=\columnwidth]{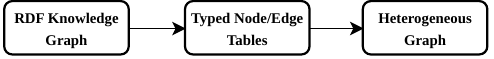}
\caption{RDF-to-heterogeneous-graph transformation process.}
\label{fig:rdf-to-graph}
\end{figure}

The resulting heterogeneous graph includes node types such as \textit{User}, \textit{Video}, \textit{InteractionContent}, \textit{TextualContent}, \textit{Language}, \textit{TargetIdentity}, \textit{PoliticalParty}, \textit{Context}, \textit{Time}, and \textit{ImpactScore}, along with edge types such as \textit{publishedBy}, \textit{belongsToVideo}, \textit{hasInteractionContext}, \textit{language}, \textit{targetTo}, \textit{publishedAt}, \textit{impactScore}, and \textit{relatedTo}. Fig.~\ref{fig:rdf-to-graph} illustrates the overall transformation process.

This transformation bridges semantic knowledge representation and graph-based computation by converting ontology-grounded RDF data into a structured heterogeneous graph while preserving node heterogeneity and relation semantics. The resulting graph serves as the foundation for constructing the perpetrator-centric subgraph and subsequent behavioral modeling.

\subsubsection{Perpetrator-Centric Subgraph Construction}
\label{subsubsec:perp-subgraph}

The heterogeneous graph constructed in Section~\ref{subsubsec:rdf-hetero-graph-transform} represents all user accounts and interactions, including non-abusive behavior. However, the objective of this work is to analyze the structural organization of user-level patterns of cyberbullying activity. To this end, we induce a perpetrator-centric subgraph by restricting the full graph to users exhibiting multiple bullying-labeled actions and their associated interaction context.

A user is defined as a perpetrator if the user has posted at least three bullying comments or replies. Let $\mathcal{U}$ denote the set of user accounts and $\mathcal{T}_{cr}(u)$ denote the set of comment and reply textual instances authored by user $u$. Since each textual instance already has a cyberbullying label assigned through the context-aware annotation process defined in Section~\ref{subsubsec:dataset}, the bullying action count of user $u\in\mathcal{U}$ is computed as

\begin{equation}
b_u =
\left\lvert
\left\{
tc \in \mathcal{T}_{cr}(u)
\mid \ell(tc)=\textit{bully}
\right\}
\right\rvert,
\label{eq:bully-count}
\end{equation}

where $b_u$ denotes the number of bullying comment or reply actions associated with user $u$. The set of perpetrators is then defined as

\begin{equation}
\mathcal{U}_p=\{u \in \mathcal{U}\mid b_u \geq \tau\},
\label{eq:perp-def}
\end{equation}

where the threshold is set to $\tau=3$. Repetition is recognized as an important characteristic of cyberbullying, although its conceptualization and operationalization vary across studies~\cite{Patchin2015Measuring}. The threshold used here does not operationalize victim-specific repetition. Instead, it serves as a study-specific user-level activity criterion for identifying users who have produced multiple bullying-labeled comments or replies across the observed interactions. Accordingly, $\tau=3$ is used to reduce the inclusion of users represented by only one or two bullying-labeled actions. This criterion is not intended as a universal definition of a cyberbullying perpetrator. Each element $p\in\mathcal{U}_p$ is subsequently treated as a perpetrator user for behavioral representation, similarity modeling, community detection, and influence analysis.

\begin{figure*}[ht]
\centering
\includegraphics[width=\textwidth]{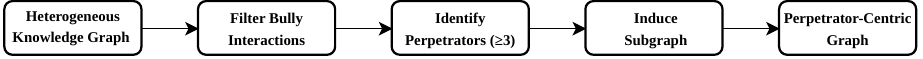}
\caption{Perpetrator-centric subgraph induction from the heterogeneous knowledge graph.}
\label{fig:perp-subgraph}
\end{figure*}

Fig.~\ref{fig:perp-subgraph} illustrates the perpetrator-centric subgraph induction process. Starting from the heterogeneous knowledge graph, bullying interactions are first filtered, users with at least $\tau=3$ bullying comments or replies are identified as perpetrators, and the corresponding interaction context is retained to induce the perpetrator-centric graph.

The subgraph is induced through a three-step filtering and retention process. First, textual content instances are filtered to retain only comments and replies whose annotation is \textit{bully}. Second, users associated with at least $\tau$ such bullying actions are identified as perpetrators. Third, a heterogeneous subgraph is constructed by retaining: (i) \textit{InteractionContent} nodes authored by perpetrators, (ii) linked \textit{TextualContent} nodes restricted to comments and replies, (iii) associated \textit{Video} nodes, and (iv) connected contextual nodes of types \textit{Language}, \textit{TargetIdentity}, \textit{Context}, \textit{Time}, \textit{ImpactScore}, and \textit{PoliticalParty}.

Although commenters and video uploaders are represented by nodes of type \textit{User} in the full heterogeneous graph, they are distinguished by role in the induced subgraph. Users who post bullying comments or replies are represented as \textit{PerpUser} nodes, whereas uploader accounts associated with videos are represented as \textit{ChannelUser} nodes. Let the resulting perpetrator-centric graph be denoted by $G_p=(\mathcal{N}_p,\mathcal{E}_p)$, where $\mathcal{N}_p$ denotes the selected \textit{PerpUser}, \textit{ChannelUser}, \textit{InteractionContent}, \textit{TextualContent}, \textit{Video}, and contextual nodes, and $\mathcal{E}_p$ denotes the typed edges connecting these nodes.

This subgraph induction process reduces irrelevant structure and concentrates the representation on behaviorally meaningful interactions. By preserving heterogeneous node and relation types while filtering non-abusive activity, the resulting graph provides a focused and semantically consistent basis for subsequent behavioral feature construction and network analysis.

\subsubsection{Behavioral and Contextual Feature Construction}
\label{subsubsec:behavior-context-feature}

The perpetrator-centric subgraph captures heterogeneous behavioral and contextual information associated with abusive users. However, similarity modeling and subsequent network analysis require a fixed-dimensional numerical representation. To address this, we construct a joint behavioral--contextual feature vector for each perpetrator through an aggregation and encoding process over the perpetrator-centric subgraph.

The feature construction process summarizes user attributes, interaction context, content exposure, engagement signals, temporal activity, and channel information into a unified representation. Aggregation is performed over all interaction-comment nodes associated with each perpetrator. Categorical attributes are accumulated as count vectors and normalized, while continuous attributes are summarized using statistical aggregation functions and discretized into comparable numerical groups.

For each perpetrator $u$, the resulting feature representation is defined as
\begin{equation}
\begin{aligned}
\mathbf{x}_p=
\big[
&\mathbf{x}_p^{(\mathrm{prof})},
\mathbf{x}_p^{(\mathrm{video})},
\mathbf{x}_p^{(\mathrm{impact})},
\mathbf{x}_p^{(\mathrm{lang})},\\
&\mathbf{x}_p^{(\mathrm{target})},
\mathbf{x}_p^{(\mathrm{party})},
\mathbf{x}_p^{(\mathrm{time})},
\mathbf{x}_p^{(\mathrm{channel})}
\big]
\in \mathbb{R}^{64},
\end{aligned}
\label{eq:feature-vector}
\end{equation}
where each component corresponds to a distinct behavioral or contextual dimension derived from the perpetrator-centric subgraph. The dimensionality of each feature group is determined from ontology-derived attributes that are consistently observable across perpetrators.

Specifically,
$\mathbf{x}_p^{(\mathrm{prof})}$ represents \emph{User Profile} features,
$\mathbf{x}_p^{(\mathrm{video})}$ represents \emph{Video} features,
$\mathbf{x}_p^{(\mathrm{impact})}$ represents \emph{Impact} features,
$\mathbf{x}_p^{(\mathrm{lang})}$ represents \emph{Language} features,
$\mathbf{x}_p^{(\mathrm{target})}$ represents \emph{Target} features,
$\mathbf{x}_p^{(\mathrm{party})}$ represents \emph{Party} features,
$\mathbf{x}_p^{(\mathrm{time})}$ represents \emph{Time} features,
and
$\mathbf{x}_p^{(\mathrm{channel})}$ represents \emph{Channel} features.

\begin{figure*}[ht]
\centering
\includegraphics[width=\textwidth]{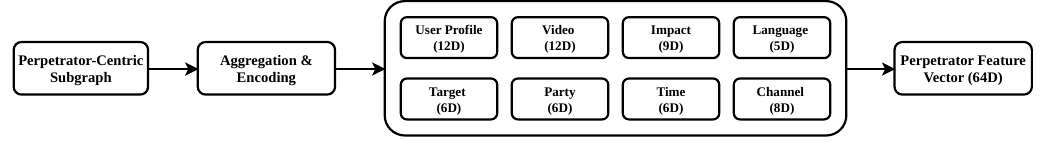}
\caption{Behavioral and contextual feature construction over the perpetrator-centric subgraph.}
\label{fig:feature-construction}
\end{figure*}

As illustrated in Fig.~\ref{fig:feature-construction}, the final representation consists of 64 features organized into eight groups: 12 user-profile features, 12 video features, 5 language features, 6 target-identity features, 6 political-context features, 6 temporal features, 9 impact features, and 8 channel-exposure features. For brevity, subsequent analyses and visualizations refer to these groups using the labels \emph{User Profile}, \emph{Video}, \emph{Impact}, \emph{Language}, \emph{Target}, \emph{Party}, \emph{Time}, and \emph{Channel}, respectively.

Categorical attributes are represented as normalized frequency distributions over observed interactions. Let $\mathbf{c}_u \in \mathbb{R}^d$ denote a categorical count vector associated with perpetrator $u$. The normalized representation is computed as

\begin{equation}
\hat{\mathbf{c}}_u
=
\frac{\mathbf{c}_u}
{\|\mathbf{c}_u\|_2+\epsilon},
\label{eq:l2-normalization}
\end{equation}

where $\epsilon$ is a small constant introduced for numerical stability.

Continuous attributes, including subscriber count, account age, content popularity, and engagement statistics, are first transformed using

\begin{equation}
x'=\log_{10}(1+x),
\label{eq:log-transform}
\end{equation}

to reduce heavy-tailed skewness commonly observed in social media data. The transformed values are subsequently encoded using a strict three-bin one-hot discretization scheme. The encoder first applies value-based tertile thresholds; if the thresholds collapse due to ties or zero inflation, a zero-aware fallback strategy separates zero-valued observations from positive values and partitions positive values using their median. This design guarantees that identical values are never assigned to different bins.

For interpretability, user-profile, video, and impact features are conceptually associated with \emph{activity intensity}, \emph{content exposure}, and \emph{engagement response}, respectively. These abstractions are used in subsequent analyses to characterize behavioral patterns at the community and network levels.

Overall, the aggregation and encoding process transforms the heterogeneous perpetrator-centric subgraph into a compact and expressive joint behavioral--contextual representation. The resulting feature space supports scalable similarity computation, interpretable behavioral comparison, and subsequent multi-scale network analysis while preserving the heterogeneous nature of cyberbullying behavior and its surrounding contextual environment.

\subsubsection{Perpetrator Similarity Modeling}
\label{subsubsec:perp-similarity-model}

To identify perpetrators with comparable behavioral signatures, we model pairwise similarity over the feature vectors defined in Eq.~\eqref{eq:feature-vector}. Let $\mathcal{P}=\mathcal{U}_p$ denote the set of perpetrator nodes. For perpetrator nodes $p,q\in\mathcal{P}$, cosine similarity is adopted because it captures directional alignment between feature vectors while reducing sensitivity to magnitude differences~\cite{Singhal2001Modern}:

\begin{equation}
\mathrm{sim}(p,q)=
\frac{\mathbf{x}_p^\top \mathbf{x}_q}
{\|\mathbf{x}_p\|_2\,\|\mathbf{x}_q\|_2},
\label{eq:cosine-sim-method}
\end{equation}

where $\mathbf{x}_p$ and $\mathbf{x}_q$ denote the behavioral representations of perpetrator nodes $p$ and $q$, respectively.

Cosine similarity is particularly suitable in this setting because the feature representation consists of heterogeneous blocks with different semantics and scales. The objective is to capture similarity in behavioral patterns rather than absolute activity magnitude, allowing perpetrators with different activity levels to be compared based on relative behavioral structure.

\textbf{Directed kNN Similarity Graph:}
Computing similarity for all pairs of perpetrators would result in a dense graph that is not suitable for network analysis. To obtain a sparse and structurally meaningful representation, we construct a directed $k$-nearest-neighbor (kNN) graph. To avoid ambiguity with the influence seed budget used later, the nearest-neighbor parameter is denoted as $k_{\mathrm{NN}}$. For each perpetrator node $p$, the top-$k_{\mathrm{NN}}$ most similar neighbors are retained, forming the set $\mathcal{N}_{k_{\mathrm{NN}}}(p)$.

This step yields a directed weighted graph, since the top-$k_{\mathrm{NN}}$ neighbor relation is not necessarily symmetric even though cosine similarity itself is symmetric. Each directed edge $(p,q)$ indicates that $q \in \mathcal{N}_{k_{\mathrm{NN}}}(p)$, and the associated edge weight is given by $\mathrm{sim}(p,q)$. The kNN construction enforces locality by preserving only the strongest behavioral relations, thereby preventing the graph from becoming fully connected.

\textbf{Symmetrized Weighted Similarity Network:}
Although cosine similarity is symmetric, the kNN construction yields a directed graph because the neighbor relation is not necessarily reciprocal. We therefore convert the directed kNN graph into an undirected weighted similarity network by averaging reciprocal edge weights when both directions exist and retaining the available weight otherwise.

The resulting similarity network is denoted as $G_s=(\mathcal{P},E_s,W_s)$, where $\mathcal{P}$ represents the set of perpetrator nodes, $E_s$ denotes the set of undirected similarity edges, and $W_s$ denotes the associated behavioral-similarity weights. This network preserves local similarity neighborhoods while ensuring compatibility with subsequent community detection, structural analysis, and diffusion modeling.

\subsection{Behavioral Community and Influence Analysis}
\label{subsec:community-influence}

This stage analyzes the perpetrator similarity network to identify multi-scale behavioral organization, interpret community-level similarity patterns, and examine influence propagation. It includes community and sub-community detection, behavioral profiling and contribution analysis, and community-constrained influence maximization.

\subsubsection{Community Detection}
\label{subsubsec:community-detect}

The perpetrator similarity network $G_s$ is expected to contain groups of perpetrator nodes with internally coherent behavioral patterns. To identify such groups, we apply modularity-based community detection using the Louvain algorithm~\cite{Blondel2008Louvain}.

The modularity objective is defined as~\cite{Newman2006Modularity}

\begin{equation}
Q=\frac{1}{2m}\sum_{p,q \in \mathcal{P}}
\left[
w_{pq}-\frac{d_p d_q}{2m}
\right]\delta(c(p),c(q)),
\label{eq:modularity-method}
\end{equation}

where $w_{pq}$ denotes the similarity weight between perpetrator nodes $p$ and $q$, $d_p=\sum_{q \in \mathcal{P}} w_{pq}$ is the weighted degree of node $p$, $m=\frac{1}{2}\sum_{p,q \in \mathcal{P}} w_{pq}$ is the total similarity weight in the network, and $\delta(c(p),c(q))$ is an indicator function equal to $1$ if $p$ and $q$ are assigned to the same community and $0$ otherwise.

The Louvain algorithm is selected due to its scalability on weighted networks and its effectiveness in identifying meso-scale structures. The resulting partition is denoted as $\mathcal{C}=\{C_1,C_2,\dots,C_K\}$, where $K$ is the number of detected communities and each $C_r \subseteq \mathcal{P}$, for $r=1,\dots,K$, represents a group of perpetrators with similar behavioral characteristics.

This step transforms pairwise behavioral similarity into an interpretable meso-scale organization of the perpetrator network. The detected communities serve as the primary structural units for subsequent behavioral profiling and community-constrained diffusion analysis.

Community detection identifies broad behavioral groups, but it may overlook variations within them. The next subsection addresses this issue by refining each community into sub-communities.

\subsubsection{Sub-community Detection}
\label{subsubsec:subcommunity-detect}

To reveal finer behavioral structures, community detection is repeated within each previously identified community. For every $C_r \in \mathcal{C}$, we construct the induced subgraph $G_s[C_r]$ and apply the Louvain algorithm independently~\cite{Blondel2008Louvain} to obtain sub-communities.

This step produces local clusters within each parent community and exposes behavioral differences that may remain hidden at the global level. A higher resolution is used for finer partitioning, while decomposition is limited to sufficiently large communities to maintain structural stability.

The resulting structure defines a two-level hierarchy in which each perpetrator node $p \in \mathcal{P}$ is associated with both a community label and a sub-community label. This hierarchical organization captures both meso-scale community-level and micro-scale sub-community-level behavioral patterns.

Sub-community detection complements the global community structure by revealing internal variations within communities, thereby supporting more fine-grained structural and behavioral analysis in subsequent sections. Each perpetrator node $p$ is thus associated with a pair $(c(p),s(p))$, where $c(p)$ denotes the community label and $s(p)$ denotes the corresponding sub-community label.

\subsubsection{Community Behavioral Profiling and Contribution Analysis}
\label{subsubsec:feat-contribute-framework}

Following community detection, we apply two complementary analyses to interpret the behavioral organization of the perpetrator network. Lift-based profiling identifies attributes that are relatively prominent within each community, while cosine-similarity decomposition quantifies how individual features and feature groups contribute to behavioral similarity.

\textbf{Lift-Based Community Profiling:}
Let $C_r\subseteq\mathcal{P}$ denote community $r$, and let $x_{p,k}$ denote the value of feature $k$ for perpetrator node $p$. The lift of feature $k$ in community $C_r$ is defined as

\begin{equation}
L_{r,k}
=
\frac{
\displaystyle
\frac{1}{\lvert C_r\rvert}
\sum_{p\in C_r}x_{p,k}
}{
\displaystyle
\frac{1}{\lvert\mathcal{P}\rvert}
\sum_{p\in\mathcal{P}}x_{p,k}
+\epsilon
},
\label{eq:feature-lift}
\end{equation}

where $\epsilon$ is a small constant used to avoid division by zero. A value of $L_{r,k}>1$ indicates that feature $k$ is over-represented in community $C_r$ relative to the overall perpetrator population.

The same calculation is applied at the feature-group level by averaging the features indexed by $\mathcal{K}_g$, where $\mathcal{K}_g$ denotes the set of features belonging to group $g$. Feature-level lift is used to identify dominant and distinctive attributes, while group-level lift summarizes the relative prominence of the eight feature groups defined in Section~\ref{subsubsec:behavior-context-feature}. For behavioral fingerprint visualization, the selected lift values are standardized column-wise across communities.

\textbf{Behavioral Contribution Analysis:}
The behavioral representation in Eq.~\eqref{eq:feature-vector} and cosine similarity in Eq.~\eqref{eq:cosine-sim-method} allow the similarity between perpetrator nodes to be decomposed into feature-wise contributions. For $p,q\in\mathcal{P}$,

\begin{equation}
\mathrm{sim}(p,q)
=
\sum_{k=1}^{64}\phi_k(p,q),
\label{eq:cosine-decomp-method}
\end{equation}

where

\begin{equation}
\phi_k(p,q)
=
\frac{x_{p,k}x_{q,k}}
{\lVert\mathbf{x}_p\rVert_2
 \lVert\mathbf{x}_q\rVert_2}
\label{eq:feature-contrib-method}
\end{equation}

denotes the contribution of feature $k$ to the similarity between $p$ and $q$.

For interpretation at the feature-group level, the feature-wise contributions are aggregated as

\begin{equation}
\Phi_g(p,q)
=
\sum_{k\in\mathcal{K}_g}\phi_k(p,q),
\label{eq:group-contrib-method}
\end{equation}

where $\mathcal{K}_g$ is the index set of features belonging to group $g$.

To characterize the contribution of feature group $g$ within community $C_r$, we average the absolute pairwise contributions over all unique perpetrator pairs in that community:

\begin{equation}
\overline{\Phi}_{g}^{(C_r)}
=
\frac{1}{\lvert\mathcal{Q}_r\rvert}
\sum_{(p,q)\in\mathcal{Q}_r}
\lvert\Phi_g(p,q)\rvert,
\label{eq:community-group-average}
\end{equation}

where

\begin{equation}
\mathcal{Q}_r
=
\{(p,q)\mid p,q\in C_r,\ p<q\}
\label{eq:community-pair-set}
\end{equation}

denotes the set of unique perpetrator pairs within community $C_r$.

To compare feature-group contributions across the full network, the group-level discriminative score is defined as

\begin{equation}
\Delta_g
=
\mathbb{E}
\left[
\lvert\Phi_g(p,q)\rvert
\mid c(p)=c(q)
\right]
-
\mathbb{E}
\left[
\lvert\Phi_g(p,q)\rvert
\mid c(p)\neq c(q)
\right],
\label{eq:delta-method}
\end{equation}

where $c(p)$ denotes the community label of perpetrator node $p$. A positive value indicates that group $g$ contributes more strongly to similarity among perpetrators in the same community than among perpetrators assigned to different communities.

The corresponding discriminative score for an individual feature is

\begin{equation}
\Delta_k
=
\mathbb{E}
\left[
\lvert\phi_k(p,q)\rvert
\mid c(p)=c(q)
\right]
-
\mathbb{E}
\left[
\lvert\phi_k(p,q)\rvert
\mid c(p)\neq c(q)
\right].
\label{eq:delta-feature-method}
\end{equation}

Lift and contribution decomposition provide different forms of interpretation. Lift identifies attributes that are relatively prominent within a community, whereas $\overline{\Phi}_{g}^{(C_r)}$, $\Delta_g$, and $\Delta_k$ measure how feature groups and individual features contribute to similarity.

\subsubsection{Community-Constrained Influence Maximization}
\label{subsubsec:im-method}

The detected communities define structurally coherent regions in which cyberbullying behavior may propagate. Rather than performing influence maximization on the full network~\cite{Kempe2003IM}, we formulate the problem independently within each community to preserve local behavioral structure and avoid dominance by large communities.

For a given community $C$, let

\begin{equation}
\mathcal{P}_C=\{p \in \mathcal{P} \mid c(p)=C\}
\label{eq:community-perp-set}
\end{equation}

denote the set of perpetrator nodes assigned to community $C$, where $c(p)$ is the community label of perpetrator node $p$.

\paragraph{Diffusion Model (Independent Cascade):}

Behavioral influence propagation is modeled using the Independent Cascade (IC) process~\cite{Kempe2003IM, Easley2010Networks}, a widely adopted diffusion model for influence analysis in social and information networks. The IC model is well suited to the proposed framework because cyberbullying interactions are modeled as stochastic peer-to-peer influence events occurring over weighted behavioral-similarity connections. Unlike threshold-based collective activation models, the IC process explicitly captures localized probabilistic transmission along network edges, enabling interpretable modeling of behavioral diffusion and efficient influence maximization.

When a perpetrator node $p$ becomes active, it has a single chance to activate each inactive neighboring perpetrator node $q$ with probability
\begin{equation}
\rho(p,q)=\min(\alpha w_{pq},\rho_{\max}),
\label{eq:edge-prob-method}
\end{equation}
where $w_{pq}$ denotes the behavioral-similarity weight between perpetrator nodes $p$ and $q$, $\alpha$ is a scaling parameter controlling diffusion strength, and $\rho_{\max}$ is an upper bound on activation probability.

To emphasize harmful users, each perpetrator node is assigned a bullying-intensity weight
\begin{equation}
\omega(p)=\log(1+b_p),
\label{eq:node-weight-method}
\end{equation}
where $b_p$ denotes the number of bullying actions performed by perpetrator $p$.

The influence of a seed set $S \subseteq \mathcal{P}_C$ is defined as the expected weighted number of activated perpetrator nodes under the IC process:
\begin{equation}
\sigma(S)=
\mathbb{E}\left[
\sum_{q \in \mathcal{A}(S)}\omega(q)
\right],
\label{eq:influence-method}
\end{equation}
where $\mathcal{A}(S)$ denotes the set of perpetrator nodes activated during diffusion.

\paragraph{Greedy Influence Maximization:}

The goal is to identify a seed set of size $k_s$ that maximizes $\sigma(S)$. Since this objective does not admit a closed-form solution, it is estimated via Monte Carlo simulation, and a greedy algorithm is used for optimization~\cite{Kempe2003IM}. At each step, the perpetrator node providing the largest marginal gain is selected:
\begin{equation}
p^*=\arg\max_{p \in \mathcal{P}_C \setminus S}
\left[
\sigma(S \cup \{p\})-\sigma(S)
\right].
\label{eq:greedy-method}
\end{equation}

This approach leverages the approximately submodular structure of the influence objective to obtain a near-optimal solution using greedy optimization~\cite{Kempe2003IM}.

\paragraph{Baseline Methods:}

To evaluate the effectiveness of the proposed approach, we consider three baseline seed-selection strategies:

\textbf{(i) Bully-Only:} Perpetrator nodes are ranked based on bullying intensity $b_p$, and the top-$k_s$ nodes are selected.

\textbf{(ii) Weighted Degree:} Perpetrator nodes are ranked based on their total similarity strength, defined as $\sum_{q \in \mathcal{N}(p)} w_{pq}$, where $\mathcal{N}(p)$ denotes the neighborhood of perpetrator node $p$.

\textbf{(iii) PageRank:} Perpetrator nodes are ranked using PageRank scores~\cite{Page1999PageRank} computed over the similarity graph, capturing global structural importance.

These baselines capture complementary aspects of influence, including individual aggressiveness, local connectivity, and global centrality.

\paragraph{Community-Constrained Formulation:}

The influence maximization procedure is applied independently within each community $C$. For each community-specific perpetrator set $\mathcal{P}_C$, a seed set of size $k_s$ is selected, and the final seed set is obtained by aggregating the selected perpetrator nodes across communities.

This formulation supports localized interventions and distributes seeds across structurally distinct behavioral groups. Section~\ref{subsec:im-analysis} evaluates the methods using both standard and bullying-weighted influence spread.

\section{Experimental Evaluation}
\label{sec:experimental-evaluation}

This section evaluates the ODCBNA framework across multiple analytical dimensions, including network structure, behavioral organization, contextual interaction, diffusion dynamics, and interpretability. The analyses are designed to address the competency questions outlined in Section~\ref{subsubsec:cb-onto}, thereby validating the effectiveness of the ontology-driven knowledge graph and the subsequent modeling pipeline.

The evaluation begins with the experimental setup, followed by multi-scale network structure analysis and community-level behavioral and contextual characterization. Subsequent analyses examine behavioral contribution, influence propagation, and diffusion dynamics, culminating in an integrated interpretation of the cyberbullying ecosystem.

\subsection{Experimental Setup}
\label{subsec:exp-setup}

This subsection outlines the evaluation network, diffusion settings, parameter-selection rationale, baseline comparison methods, and measures used in the experiments.

\textbf{Network Construction and Scope:}
The experiments use the perpetrator similarity network $G_s$ constructed from the behavioral representations described in Section~\ref{subsubsec:behavior-context-feature}. It contains $|\mathcal{P}|=1{,}063$ perpetrator nodes connected by weighted similarity edges obtained through a $k$-nearest-neighbor strategy. To distinguish the neighborhood size from the influence seed budget, we denote it by $k_{\mathrm{NN}}$ and set $k_{\mathrm{NN}}=40$. The directed kNN graph is subsequently symmetrized to obtain the undirected weighted network $G_s$.

\textbf{Diffusion Parameters:}
Influence propagation follows the Independent Cascade (IC) process introduced in Section~\ref{subsubsec:im-method}. We use a seed budget of $k_s=3$, diffusion-strength parameter $\alpha=0.5$, upper bound on the activation probability $\rho_{\max}=0.25$, $R_{\mathrm{sel}}=120$ Monte Carlo runs during seed selection, and $R_{\mathrm{eval}}=300$ runs for final influence evaluation.

\textbf{Parameter Selection Rationale:}
The adopted parameter values are treated as study-specific experimental settings rather than universally optimal choices. For network construction, $k_{\mathrm{NN}}=40$ is selected to maintain a connected similarity network while preserving locality by restricting each perpetrator to a limited set of behaviorally similar neighbors. The seed budget $k_s=3$ represents a constrained intervention in which only a small number of perpetrator nodes can be selected within each community. In the IC model, $\alpha$ scales the behavioral-similarity weights used to determine activation probabilities, while $\rho_{\max}$ bounds these probabilities to avoid excessively large activation likelihoods. Monte Carlo simulation follows established practice~\cite{Kempe2003IM}. A smaller Monte Carlo budget, $R_{\mathrm{sel}}$, is used during greedy seed selection because influence spread must be estimated repeatedly across candidate nodes, whereas the larger budget, $R_{\mathrm{eval}}$, is used for final evaluation to obtain more reliable spread estimates. All seed-selection methods are evaluated using the same diffusion parameters and seed budget to ensure comparability. These settings define the operating point of the present experiments rather than a claim of parameter optimality.

\textbf{Baseline Methods:}
We compare the greedy influence maximization strategy with three baseline seed-selection strategies introduced in Section~\ref{subsubsec:im-method}, namely Bully-Only, Weighted Degree, and PageRank~\cite{Page1999PageRank}.

\textbf{Evaluation Measures:}
Different measures are used for each analytical task. Network structure is examined through density, average degree, clustering coefficient, average shortest path length, and modularity. Community profiles are described using dominant attributes and lift-based distinctive features. Influence maximization is evaluated through expected spread, relative gain over the strongest baseline, and overlap among selected seed sets. Feature interpretability is assessed by comparing contribution values for within-community and between-community perpetrator pairs at both group and individual feature levels.

\subsection{Multi-scale Network Structure}
\label{subsec:network-analysis}

This subsection examines the perpetrator similarity network at three structural scales: the global network, communities, and sub-communities. It characterizes network connectivity and the hierarchical organization of similarity-based behavioral groupings.

\textbf{Global Network Structure:}
Table~\ref{tab:network-overview} summarizes the perpetrator, community, and sub-community networks. The perpetrator similarity network contains $1{,}063$ nodes and $27{,}554$ undirected edges, with an average degree of $51.84$ and a density of $0.0488$. This low density reflects the sparse connectivity produced by the $k$-nearest-neighbor construction.

\begin{table}[width=\linewidth,cols=4,pos=h]
\centering
\caption{Structural properties of the perpetrator, community, and sub-community networks.}
\label{tab:network-overview}
\footnotesize
\begin{tabular*}{\tblwidth}{@{} LCCCCCC@{} }
\toprule
\textbf{Network} & \textbf{\#Nodes} & \textbf{\#Edges} & \textbf{Avg Deg} & \textbf{Density} & \textbf{Clust} & \textbf{Path} \\
\midrule
Perpetrator   & 1063 & 27554 & 51.84 & 0.0488 & 0.1939 & 3.4192 \\
Community     & 6    & 15    & 5.00  & 1.0000 & 1.0000 & 0.0015 \\
Sub-community & 27   & 311   & 23.04 & 0.8860 & 0.8932 & 0.0156 \\
\bottomrule
\end{tabular*}

\vspace{2pt}
\parbox{\columnwidth}{\footnotesize
\textbf{Notes:} Avg Deg = Average degree; Clust = Average clustering coefficient; Path = Average weighted shortest-path distance in the largest connected component.
}
\end{table}

The perpetrator network has an average clustering coefficient of $0.1939$ and an average weighted shortest-path distance of $3.4192$. These values indicate some local grouping and relatively short paths within the largest connected component, consistent with characteristics associated with small-world networks~\cite{Watts1998SmallWorld}. A comparison with an appropriate random-network baseline would be required for formal confirmation.

At the aggregated levels, the community network is complete, while the sub-community network remains highly connected, with a density of $0.8860$ and a clustering coefficient of $0.8932$.

Interactive visualizations of the perpetrator similarity network are available at~\url{https://cyberbly-na.vercel.app/visualizations/perpetrator-similarity}.

\textbf{Community-Level Structure:}
The Louvain resolution parameter was selected by examining candidate values between $0.70$ and $1.20$. Among the tested values, $0.85$ yielded the highest modularity of approximately $0.495$, identifying six communities with sizes ranging from $104$ to $237$ perpetrators. Table~\ref{tab:community-structure} reports the internal structural properties and meta-level connectivity of each community.

\begin{table}[cols=4,pos=h]
\centering
\caption{Community-level structural properties of the perpetrator network.}
\label{tab:community-structure}
\footnotesize
\begin{tabular*}{\tblwidth}{@{} CCCCCCC@{} }
\toprule
\textbf{Comm.} & \textbf{\#Nodes} & \textbf{\#Edges} & \textbf{Density} & \textbf{Avg Deg} & \textbf{Clust} & \textbf{Meta Deg} \\
\midrule
C0 & 213 & 3214 & 0.1424 & 30.18 & 0.2676 & 5 \\
C1 & 141 & 2135 & 0.2163 & 30.28 & 0.2984 & 5 \\
C2 & 184 & 3728 & 0.2214 & 40.52 & 0.3208 & 5 \\
C3 & 104 & 1514 & 0.2827 & 29.12 & 0.3621 & 5 \\
C4 & 237 & 4010 & 0.1434 & 33.84 & 0.2570 & 5 \\
C5 & 184 & 3556 & 0.2112 & 38.65 & 0.3157 & 5 \\
\bottomrule
\end{tabular*}

\vspace{2pt}
\parbox{\columnwidth}{\footnotesize
\textbf{Notes:} Avg Deg = Average internal degree; Clust = Average weighted clustering coefficient; Meta Deg = Degree in the community meta-network.
}
\end{table}

C3, the smallest community, has the highest internal density of $0.2827$, whereas C0 and C4, the two largest communities, have the lowest densities of $0.1424$ and $0.1434$, respectively. This indicates that the smaller community contains a greater proportion of possible internal links, while the two largest communities exhibit more dispersed connectivity.

Average degree follows a different pattern. C2 has the highest value of $40.52$, followed by C5 at $38.65$, whereas C3 has the lowest at $29.12$. Weighted clustering ranges from $0.2570$ to $0.3621$, with C3 showing the strongest local clustering and C4 the weakest. These variations indicate that community size, internal connectivity, and local clustering capture different aspects of the network structure.

All communities have a meta-degree of $5$, meaning that each connects to every other community in the meta-network. They therefore differ internally but remain interconnected at the aggregated level.

Interactive visualizations of the community network are available at~\url{https://cyberbly-na.vercel.app/visualizations/perpetrator-community}.

\textbf{Sub-community (Hierarchical) Structure:}
Hierarchical Louvain partitioning was applied within the six parent communities to examine finer-grained behavioral groupings. Similar to community detection, candidate resolution values between $0.70$ and $1.20$ were examined, with $1.02$ selected for sub-community detection. At this resolution, the analysis identifies $27$ sub-communities, with an edge-weighted average modularity of approximately $0.314$. This value represents the average of the within-parent modularity scores, weighted by the number of internal edges in each parent community. Table~\ref{tab:subcommunity-structure} summarizes the sub-community distribution within each parent community.

\begin{table}[width=.9\linewidth,cols=4,pos=h]
\centering
\caption{Hierarchical sub-community structure within the six parent communities.}
\label{tab:subcommunity-structure}
\footnotesize
\begin{tabular*}{\tblwidth}{@{} CCCC@{} }
\toprule
\textbf{Parent} & \textbf{\#Subcomms} & \textbf{Avg Size} & \textbf{Largest Size} \\
\midrule
C0 & 4 & 53.25 & 64 \\
C1 & 4 & 35.25 & 51 \\
C2 & 6 & 30.67 & 42 \\
C3 & 4 & 26.00 & 38 \\
C4 & 5 & 47.40 & 63 \\
C5 & 4 & 46.00 & 57 \\
\bottomrule
\end{tabular*}
\end{table}

The number of sub-communities varies from four to six per parent community. C2 contains the highest number of sub-communities (six), followed by C4 with five, while the remaining communities each contain four. C0 has the largest average sub-community size ($53.25$), whereas C3 has the smallest ($26.00$). The largest individual sub-communities occur in C0 and C4, containing $64$ and $63$ perpetrators, respectively. These differences indicate that the parent communities exhibit varying levels of internal subdivision, revealing finer-grained groupings within the broader behavioral communities.

Interactive visualizations of the sub-community network are available at~\url{https://cyberbly-na.vercel.app/visualizations/perpetrator-subcommunity}.

Overall, the network of $1{,}063$ perpetrators exhibits a hierarchical structure comprising six communities and $27$ sub-communities. The community and sub-community partitions reveal progressively finer groupings within the broader similarity network.

\subsection{Community Behavior}
\label{subsec:community-behavior}

This subsection characterizes the six communities using condensed profiles, distinctive attributes, and behavioral fingerprints. The analysis considers the eight feature groups introduced in Section~\ref{subsubsec:behavior-context-feature}: User Profile, Video, Impact, Language, Target, Party, Time, and Channel.

\textbf{Community-Level Behavior:}
Table~\ref{tab:community-profile} summarizes community size, interaction activity, bullying activity, and the most prominent attributes from each feature group. C4 and C1 record more than 5.4 actions per perpetrator, followed by C0 with 5.16. The corresponding values for C2, C3, and C5 remain close to 4.0. C4 is both the largest and the most active community, containing 237 perpetrators and 1,351 interaction actions. It also has the highest number of bullying actions per perpetrator at 4.46, followed by C1 at 4.36.

\begin{table*}[cols=4,pos=h]
\centering
\caption{Condensed community-level behavioral profiles using dominant attributes from each feature group.}
\label{tab:community-profile}
\footnotesize
\setlength{\tabcolsep}{3pt}

\resizebox{\textwidth}{!}{%
\begin{tabular}{c c c c c c l l l l l l l l}
\textbf{Comm.} & \textbf{Perp.} & \textbf{Act.} & \textbf{Act./P} & \textbf{Bully} & \textbf{B/P} & \textbf{Language} & \textbf{Target} & \textbf{Party} & \textbf{Time} & \textbf{Channel} & \textbf{User Profile} & \textbf{Video} & \textbf{Impact} \\
\midrule

C0 & 213 & 1099 & 5.16 & 864 & 4.06 & Bengali & Individual / & AL / & Afternoon / & Rtv News / & High Video / & High Avg. Comment / & High \% Reply / \\
& & & & & & & General People & Other & Morning & Ekattor TV & High View & High Avg. Like & High \% Like \\

C1 & 141 & 774 & 5.49 & 615 & 4.36 & -- & People Group / & NCP / & Night / & DBC News / & Low Acc. Age / & Low Avg. View / & Mid p95 Likes / \\
& & & & & & & Male & JaPa & Late Night & Channel 24 & Low View & Low Avg. Like & High \% Reply \\

C2 & 184 & 749 & 4.07 & 663 & 3.60 & Romanized & Organization & JaPa & Early Morning / & DBC News / & Low Subscribe / & Low Avg. Like / & Low p95 Likes / \\
& & & & & & Bengali & & & Late Night & Independent TV & Low Video & Low Avg. View & Low \% Like \\

C3 & 104 & 441 & 4.24 & 387 & 3.72 & Romanized & Organization / & JaPa / & Late Night / & Independent TV / & Low Acc. Age / & Mid Avg. Like / & Low \% Like / \\
& & & & & & Bengali & Female & Other & Morning & Ekattor TV & Low View & Mid Avg. View & Low p95 Likes \\

C4 & 237 & 1351 & 5.70 & 1058 & 4.46 & English & Female / & Jamaat / & Evening / & Jamuna TV / & Low Video / & High Avg. Like / & Mid \% Like / \\
& & & & & & & Male & BNP & Early Morning & Somoy TV & Low View & Mid Avg. View & Mid \% Reply \\

C5 & 184 & 745 & 4.05 & 684 & 3.72 & Romanized & Female / & AL / & Night / & Jamuna TV / & Low Subscribe / & High Avg. View / & Low \% Like / \\
& & & & & & Bengali & People Group & Jamaat & Evening & Somoy TV & Low Video & High Avg. Like & Low p95 Likes \\

\bottomrule
\end{tabular}%
}

\vspace{2pt}
\parbox{\textwidth}{\footnotesize
\textbf{Notes:} Comm. = Community; Perp. = Number of perpetrators; Act. = Total interaction actions; Act./P = Actions per perpetrator; Bully = Bullying actions; B/P = Bullying actions per perpetrator.
Each feature-group column reports at most the two attributes with the largest lift values for the corresponding community. -- indicates that no dominant attribute was reported for that group.
}
\end{table*}

C2 and C3 display similar contextual patterns, including Romanized Bengali, organization-focused targets, and JaPa-related discussions. Both are associated with lower user-profile activity attributes, although their video characteristics differ. C2 is linked to low average video views and likes, whereas C3 shows mid-range values. C0 and C1 exhibit higher interaction activity but distinct behavioral profiles. C0 is associated with high user video and view counts, together with stronger reply- and like-based impact attributes. In contrast, C1 is characterized by lower user-profile and video engagement attributes.

C4 combines the highest interaction activity with English content, female and male targets, Jamaat- and BNP-related discussions, and comparatively stronger mid-range impact attributes. C5 is associated with Romanized Bengali, female and people-group targets, AL- and Jamaat-related discussions, and high average video views and likes. However, its interaction-level impact attributes remain predominantly low. This contrast indicates that greater video engagement does not necessarily correspond to stronger responses to individual interactions.

\textbf{Distinctive Attributes:}
Table~\ref{tab:distinctive-attributes} lists the three attributes with the largest lift values in each community. Lift, defined in Eq.~\eqref{eq:feature-lift}, compares the mean value of an attribute within a community with its mean across all perpetrators. Values greater than $1$ indicate relative over-representation.

\begin{table}[ht]
\centering
\caption{Top three distinctive attributes of each community based on lift.}
\label{tab:distinctive-attributes}
\footnotesize
\setlength{\tabcolsep}{4pt}
\begin{tabular}{c p{6.8cm}}
\toprule
\textbf{Comm.} & \textbf{Top Distinctive Attributes} \\
\midrule
C0 & High Video (4.64), High View (4.55), Mid Video (2.77) \\
C1 & Low Avg. View (2.61), Low Avg. Like (2.44), Mid p95 Likes (2.22) \\
C2 & Low Avg. Like (2.75), Low Avg. View (2.59), Low p95 Likes (2.02) \\
C3 & Mid Avg. Like (3.00), Mid Avg. View (2.74), Low \% Like (1.93) \\
C4 & Mid \% Like (2.35), Mid \% Reply (2.10), Mid p95 Likes (2.05) \\
C5 & High Avg. View (2.61), High Avg. Like (2.53), Low \% Like (2.03) \\
\bottomrule
\end{tabular}
\end{table}

Most of the leading attributes belong to the User Profile, Video, and Impact groups. C1 and C2 are distinguished by low average video views and likes, with additional differences in peak-like activity. C3, in contrast, is characterized by mid-range average views and likes, together with a low proportion of liked interactions. C5 combines high average video views and likes with a low proportion of liked interactions.

The strongest lift values occur in C0 and relate to User Profile, particularly high video and view counts. C4 is differentiated mainly by interaction-response attributes, including mid-range liked and replied proportions and peak likes. These patterns indicate that community distinctions arise from different combinations of account activity, video exposure, and engagement response.

\begin{figure*}[ht]
\centering
\includegraphics[width=\textwidth]{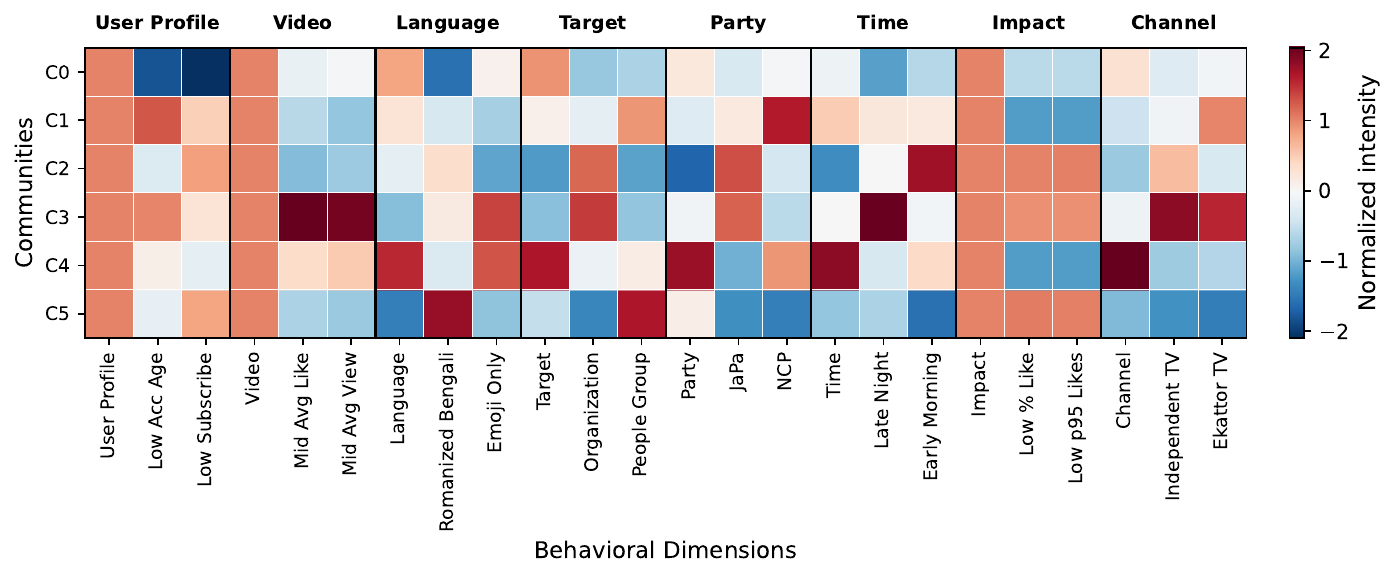}
\caption{Community behavioral fingerprints based on column-wise standardized lift values.}
\label{fig:community-fingerprint}
\end{figure*}

\textbf{Behavioral Fingerprint Analysis:}
Fig.~\ref{fig:community-fingerprint} provides a compact comparison of the community profiles. It includes the aggregate profile of each feature group together with up to two representative attributes from that group. The selected lift values are standardized column-wise across communities before plotting. The colors therefore represent relative variation within each dimension rather than differences in their original scales.

The fingerprints reveal several contrasts. C3 shows strong positive intensity in mid-range average video likes and views, whereas C2 and C5 exhibit weaker signals for these attributes. C4 has stronger aggregate Language, Target, Party, Time, and Channel signals. Language-related differences include Romanized Bengali in C5 and emoji-only content in C3. Target-related variation is particularly visible in organization-focused activity in C2 and people-group associations in C5.

Additional variation appears across Party, Time, and Channel. C1 is associated with NCP-related discussions, while C2 shows stronger JaPa and early-morning signals. C3 is distinguished by late-night activity and associations with Independent TV and Ekattor TV. C4 exhibits relatively strong aggregate Party, Time, and Channel intensities, indicating broader contextual differences among the communities.

Impact patterns also separate the communities. Low liked-interaction and low peak-like attributes are more prominent in C2, C3, and C5, but less prominent in C1 and C4. These differences reflect the relative representation of specific engagement-response categories rather than absolute interaction intensity. Taken together, the fingerprints reveal variation in video engagement, interaction response, language use, target identity, political context, temporal activity, and channel exposure.

\subsection{Community Interaction}
\label{subsec:community-interaction}

To complement the behavioral profiles in Section~\ref{subsec:community-behavior}, we construct two aggregated views: a \textit{Community--Behavioral Attribute Interaction Network} and a \textit{Community--Context Interaction Network}. These networks summarize community associations with behavioral and contextual attributes. Edge weights represent perpetrator counts in the behavioral network and interaction counts in the contextual network, while edge intensity indicates the corresponding bullying proportion.

\textbf{Community--Behavioral Attribute Interaction Network:}
The network connects communities with attributes from the User Profile, Video, and Impact groups. An interactive visualization is available at~\url{https://cyberbly-na.vercel.app/visualizations/community-behavior}.

For the User Profile group, high subscriber count is prominent in C0, whereas low view and video counts dominate several other communities. However, the highest bullying proportions occur in different categories, including low subscriber count in C0, mid view count in C2--C4, and high video count in C5.

The Video group reveals differences in content exposure and engagement. High average comment count is prominent in C0, low average views or likes in C1 and C2, mid average likes in C3, and high average likes or views in C4 and C5. The highest bullying proportions often correspond to different attributes, such as low video duration in C0, mid average views in C2, and low average views in C4.

For the Impact group, low percentage of replied interactions is prominent in C0, C1, C3, and C4, while low peak-like count is prominent in C2 and C5. In contrast, high percentage of replied interactions records the highest bullying proportion in C0, C1, and C4, whereas low percentage of liked interactions is associated with the highest proportion in C2.

Overall, attributes associated with more perpetrators do not necessarily exhibit higher bullying proportions, highlighting differences in behavioral patterns among communities.

\textbf{Community--Context Interaction Network:}
The network connects communities with attributes from the Language, Target, Party, Time, and Channel groups, comparing interaction volume and bullying proportion. An interactive visualization is available at~\url{https://cyberbly-na.vercel.app/visualizations/community-context}.

For the Language group, bengali accounts for the greatest interaction volume in all communities. However, romanized bengali records the highest bullying proportion in C0, C2, and C4, while code-mixed content leads in C5, with all $26$ interactions labeled as bullying.

The Target group shows that male targets dominate interaction volume in C0, C1, C4, and C5, whereas organization-focused interactions lead in C2 and C3. Higher bullying proportions are associated with female targets in C0, people groups in C1, C2, and C4, general people in C3, and organizations in C5.

For the Party group, NCP accounts for the greatest interaction volume in C0, C1, and C4, JaPa in C2 and C3, and Jamaat in C5. In contrast, AL records the highest bullying proportion in four communities, while JaPa leads in C2 and C5.

For the Time group, afternoon dominates interaction volume in every community, whereas the highest bullying proportions occur mainly at night, except in C2 (early morning) and C5 (evening).

The Channel group reveals further differences. Rtv News leads interaction volume in C0, Channel 24 in C1, Independent Television in C2 and C3, and Jamuna TV in C4 and C5. However, Jamuna TV records the highest bullying proportion in C0, C1, C2, and C4, followed by DBC NEWS in C3 and Ekattor TV in C5.

Overall, greater interaction volume does not necessarily correspond to higher bullying proportions, highlighting contextual differences in community activity and bullying-labeled interactions.

\subsection{Behavioral Contribution Analysis}
\label{subsec:feature-contribution-analysis}

The preceding analyses describe the structural organization and behavioral patterns of perpetrators, but they do not show which dimensions contribute most to the observed similarity structure. We therefore examine feature contributions at the group, individual, and community levels using the framework defined in Section~\ref{subsubsec:feat-contribute-framework}.

\begin{figure}
\centering
\includegraphics[width=\columnwidth]{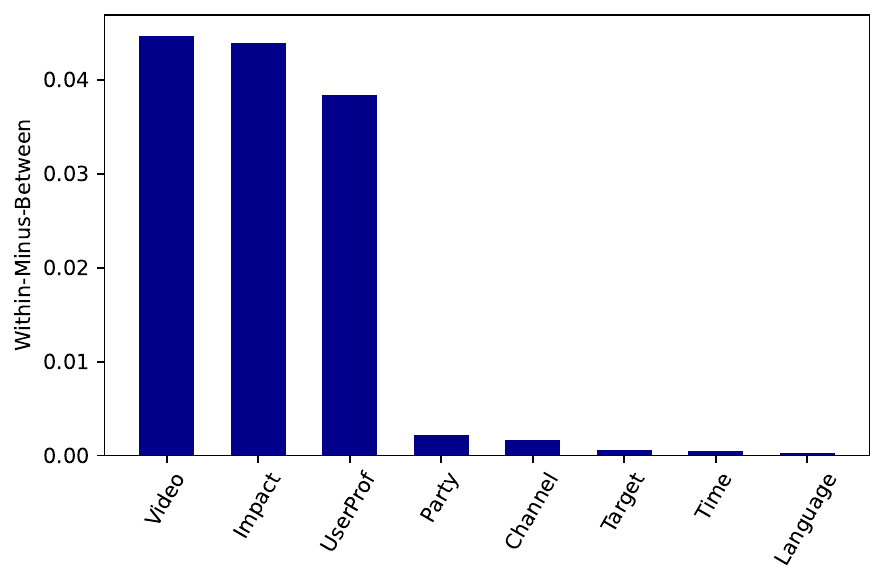}
\caption{Group-level differences between within-community and between-community similarity contributions.}
\label{fig:group-contribution}
\end{figure}

\textbf{Group-Level Contribution Analysis:}
Fig.~\ref{fig:group-contribution} reports the group-level discriminative scores computed using Eq.~\eqref{eq:delta-method}. Each score represents the difference between the mean contribution of a feature group for within-community pairs and that for between-community pairs. A larger positive value indicates a stronger contribution to within-community similarity relative to between-community similarity.

The Video group records the largest discriminative contribution, followed closely by the Impact group and then the User Profile group. Their scores are substantially greater than those of the Party, Channel, Target, Time, and Language groups, whose contributions remain comparatively small.

These results indicate that community separation is associated mainly with video exposure, engagement response, and user-profile characteristics. The contextual feature groups contribute less directly to distinguishing within-community similarity from between-community similarity.

\begin{figure}
\centering
\includegraphics[width=\columnwidth]{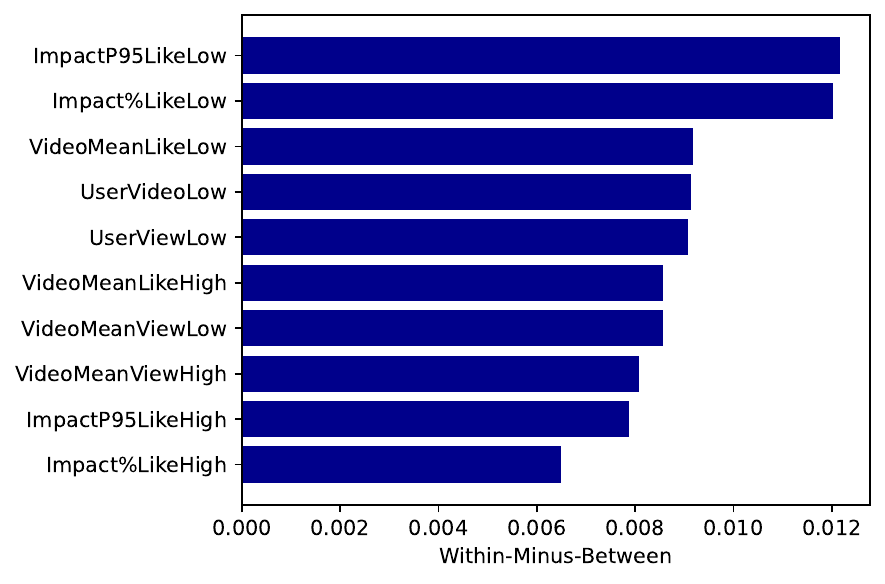}
\caption{Top 10 individual features based on within-community and between-community contribution differences.}
\label{fig:individual-contribution}
\end{figure}

\textbf{Individual-Level Contribution:}
Fig.~\ref{fig:individual-contribution} presents the 10 features with the largest discriminative scores computed using Eq.~\eqref{eq:delta-feature-method}. The ranking considers all 64 features, while only the leading subset is shown for readability. A larger positive score indicates a stronger contribution to within-community similarity relative to between-community similarity.

The leading features come entirely from the Impact, Video, and User Profile groups. The strongest scores are associated with low peak-like count and low percentage of liked interactions, followed by low average video likes. Low account-level video and view counts also rank among the leading attributes. These results indicate that lower engagement response, video exposure, and user activity contribute prominently to distinguishing within-community similarity from between-community similarity.

The remaining leading features include high average video likes, low and high average video views, high peak-like count, and high percentage of liked interactions. Together, they indicate that both lower and higher ranges of video exposure and engagement response contribute to community differentiation.

The Language, Target, Party, Time, and Channel groups are absent from the displayed subset, indicating smaller individual discriminative contributions than the leading behavioral features.

\begin{figure}
\centering
\includegraphics[width=\columnwidth]{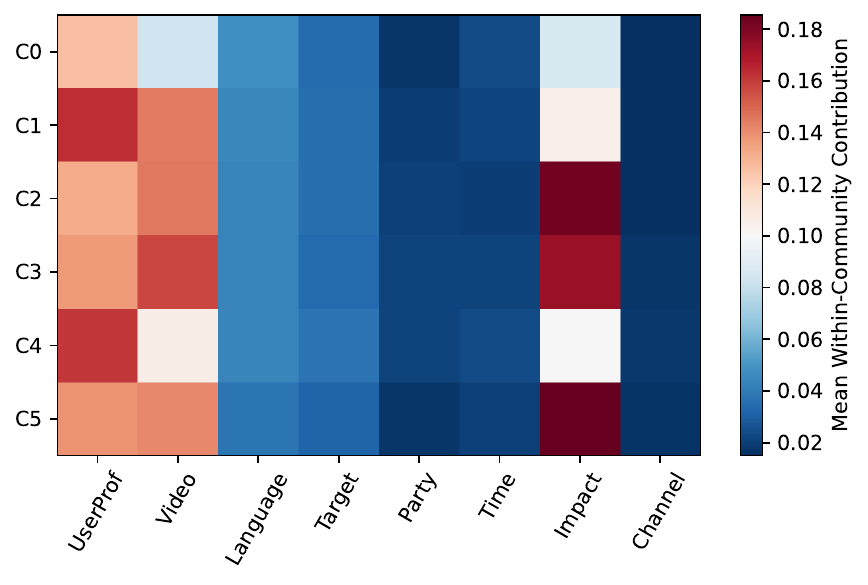}
\caption{Community-wise mean contributions of behavioral and contextual feature groups to within-community similarity.}
\label{fig:community-group-contribution}
\end{figure}

\textbf{Community-Specific Contribution Patterns:}
Fig.~\ref{fig:community-group-contribution} presents the mean within-community contribution of each feature group. Pairwise group contributions are computed using Eq.~\eqref{eq:group-contrib-method} and averaged within each community according to Eq.~\eqref{eq:community-group-average}.

The User Profile group provides the largest contribution in C0, C1, and C4. In C0, it is followed by the Impact and Video groups, whereas the Video group ranks second in C1 and C4. In contrast, the Impact group dominates C2, C3, and C5, followed by the Video and User Profile groups.

In all six communities, the User Profile, Video, and Impact groups contribute substantially more than the Language, Target, Party, Time, and Channel groups. Their relative prominence varies among communities, indicating that within-community similarity reflects different combinations of account characteristics, video exposure, and engagement response.

\subsection{Influence Propagation}
\label{subsec:im-analysis}

This subsection evaluates the community-constrained influence maximization framework from three perspectives: spread performance, seed-selection behavior, and diffusion dynamics.

\begin{figure}
\centering
\includegraphics[width=\columnwidth]{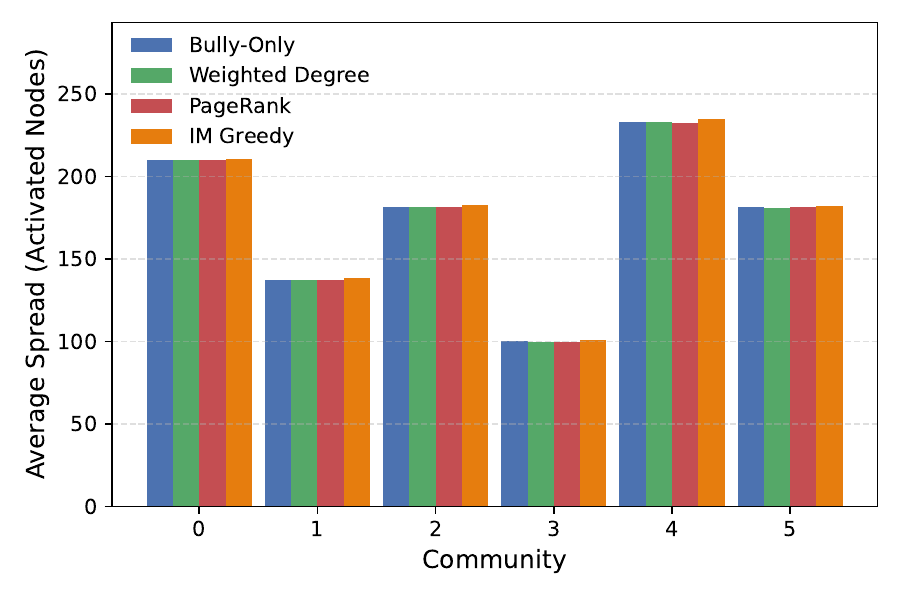}
\caption{Average influence spread of the seed-selection methods across communities for a seed budget of $k_s=3$.}
\label{fig:spread-comparison}
\end{figure}

\textbf{Influence Spread Performance:}
Fig.~\ref{fig:spread-comparison} compares the average number of activated nodes achieved by the greedy method and the three baseline strategies in the six communities.

The four methods produce closely comparable spreads in every community. The greedy method achieves marginally higher spread than the baselines in all six communities, although the differences remain small. C4 records the highest overall spread, followed by C0, while C3 exhibits the lowest. The differences among methods are considerably smaller than those among communities, indicating that distinct seed-selection strategies achieve similar diffusion reach under the adopted IC parameters and seed budget.

\begin{figure}
\centering
\includegraphics[width=\columnwidth]{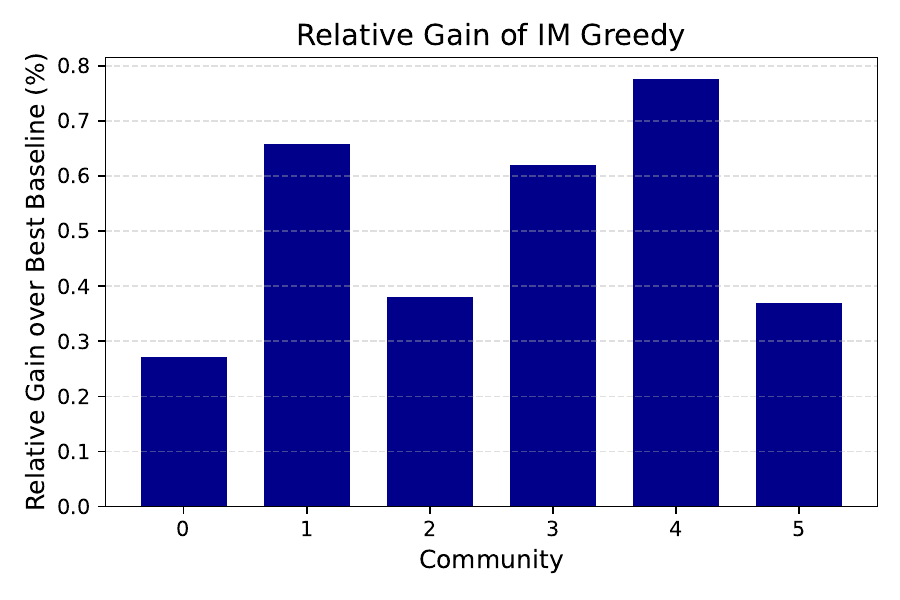}
\caption{Percentage difference between the greedy method and the strongest baseline across communities.}
\label{fig:percent-gain}
\end{figure}

\textbf{Relative Gain Analysis:}
Fig.~\ref{fig:percent-gain} reports the percentage difference between the greedy method and the strongest baseline in each community.

The greedy method records positive gains in all six communities. The improvement is largest in C4 at approximately 0.77\%, followed by C1 at 0.66\% and C3 at 0.62\%. Gains in C2 and C5 remain below 0.4\%, while C0 records the smallest improvement at approximately 0.27\%.

Overall, the baseline strategies already achieve spreads close to those obtained by the greedy method. Direct optimization provides consistent but modest improvements across the six communities under the adopted IC settings.

\textbf{Seed Selection Divergence:}
Table~\ref{tab:seed-overlap} compares the greedy seed sets with those selected by the three baseline strategies for $k_s=3$.

\begin{table}[width=\linewidth,cols=4,pos=h]
\centering
\caption{Seed overlap between the greedy method and baseline strategies for $k_s=3$.}
\label{tab:seed-overlap}
\footnotesize
\begin{tabular*}{\tblwidth}{@{} CCCCCCC@{} }
\toprule
\textbf{Comm.} & \textbf{IM$\cap$B} & \textbf{IM$\cap$WD} & \textbf{IM$\cap$PR} & \textbf{B (\%)} &
\textbf{WD (\%)} & \textbf{PR (\%)} \\
\midrule
C0 & 0 & 0 & 0 & 0.00  & 0.00 & 0.00 \\
C1 & 0 & 0 & 0 & 0.00  & 0.00 & 0.00 \\
C2 & 1 & 0 & 0 & 33.33 & 0.00 & 0.00 \\
C3 & 1 & 0 & 0 & 33.33 & 0.00 & 0.00 \\
C4 & 0 & 0 & 0 & 0.00  & 0.00 & 0.00 \\
C5 & 0 & 0 & 0 & 0.00  & 0.00 & 0.00 \\
\bottomrule
\end{tabular*}

\vspace{2pt}
\parbox{\columnwidth}{\footnotesize
\textbf{Notes:} Comm. = Community; IM = Greedy influence maximization;
B = Bully-Only; WD = Weighted Degree; PR = PageRank.
Overlap percentages are calculated relative to the seed budget.
}
\end{table}

The greedy seed sets have limited overlap with those of the baselines. In C2 and C3, the greedy and Bully-Only methods share one seed each, corresponding to 33.33\% of the three-node seed set. No overlap with Weighted Degree or PageRank appears in any community. An additional evaluation with $k_s=5$ shows Bully-Only overlap of 20\% in C2 and 40\% in C3, with no overlap elsewhere or with the other baselines.

Despite selecting largely different nodes, the methods achieve closely comparable influence spreads for $k_s=3$. This indicates that distinct seed configurations can provide similar diffusion reach under the adopted IC setting, suggesting that influence propagation is not dependent on a single seed configuration.

\begin{figure*}[ht]
\centering
\includegraphics[width=\textwidth]{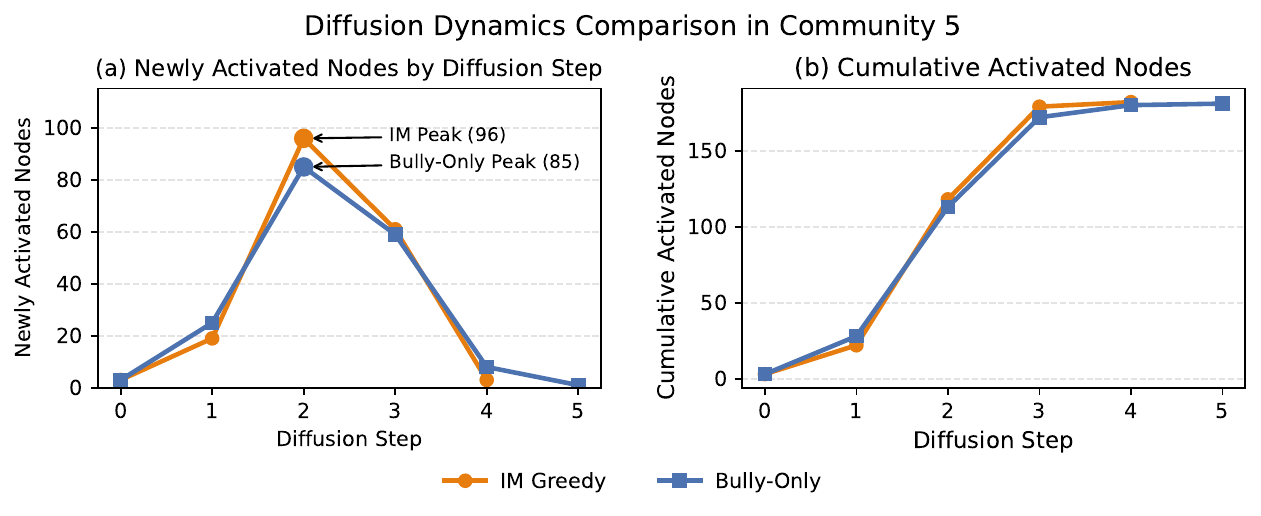}
\caption{Representative diffusion dynamics of IM Greedy and Bully-Only in community C5.}
\label{fig:diffusion-dynamics}
\end{figure*}

\textbf{Diffusion Dynamics:}
Diffusion dynamics were examined for all six communities. As an illustrative example, Fig.~\ref{fig:diffusion-dynamics} presents C5, comparing the representative diffusion trajectories of the greedy method and the Bully-Only baseline.

Both methods exhibit similar activation patterns, with newly activated nodes peaking at step 2. The Bully-Only baseline activates more nodes at step 1, whereas the greedy method reaches a higher peak at step 2, activating 96 nodes compared with 85 for the baseline. Activation subsequently declines for both methods. The greedy method reaches a diffusion depth of four steps, while the baseline continues for five steps.

The cumulative curves remain close throughout the process. The baseline leads initially, but the greedy method overtakes it at step 2 and finishes with 182 activated nodes compared with 181 for Bully-Only. Their average spreads are also comparable, at 181.84 and 181.17 activated nodes, respectively. These results indicate that different seed configurations can achieve similar diffusion reach while exhibiting differences in activation timing and diffusion depth.

Visualizations of influence propagation in all six communities under the four seed-selection methods are available at~\url{https://cyberbly-na.vercel.app/visualizations/diffusion-all-methods}.

Together with the limited seed overlap, the similar diffusion trajectories observed in C5 suggest that distinct seed configurations can achieve comparable diffusion reach under the adopted IC setting.

\subsection{Discussion}
\label{subsec:discussion}

The findings reported in Sections~\ref{subsec:network-analysis}--\ref{subsec:im-analysis} are discussed collectively to explain how network structure, behavioral similarity, contextual variation, and influence propagation characterize perpetrator behavior within the ODCBNA framework.

\textbf{Behavioral Organization:}
The six communities exhibit differences in size, internal connectivity, and behavioral profiles, with further variation revealed by the $27$ sub-communities. Community-level profiles and fingerprints show differences in user characteristics, video exposure, and engagement response, with the prominence of these dimensions varying among communities. These findings indicate that perpetrators with similar behavioral patterns form distinguishable groups while retaining internal structural and behavioral variation.

\textbf{Contextual Embedding:}
The Language, Target, Party, Time, and Channel groups contribute relatively little to behavioral similarity and community separation, but help characterize the settings in which bullying-labeled interactions occur. The community--context interaction network further shows that attributes associated with greater interaction volume do not necessarily exhibit higher bullying proportions. Thus, behavioral similarity and contextual variation provide different but related perspectives on perpetrator activity and community characteristics.

\textbf{Behavioral Sources of Similarity:}
At the group level, the Video group records the largest discriminative contribution, followed closely by the Impact and User Profile groups. The leading individual features also belong entirely to these three groups, while the contextual groups have substantially smaller discriminative scores. Community-specific analysis reveals variation in their relative contributions, with the User Profile group most prominent in C0, C1, and C4, and the Impact group in C2, C3, and C5. These findings indicate that behavioral similarity reflects different combinations of video exposure, engagement response, and account-level characteristics rather than any single feature group or contextual attribute.

\textbf{Influence Propagation:}
The greedy method achieves small positive gains in all six communities, although the strongest baselines remain competitive. The seed sets show limited overlap, with only one shared seed between IM Greedy and Bully-Only in each of C2 and C3. The illustrative analysis of C5 further shows similar diffusion trajectories despite differences in peak activation and diffusion depth. These findings suggest that distinct seed configurations can achieve comparable diffusion reach under the adopted IC parameters.

\textbf{Overall Interpretation:}
The findings reveal a multidimensional system linking network structure, behavioral similarity, contextual variation, and simulated influence propagation. Communities reflect different combinations of user characteristics, video exposure, and engagement response, while contextual analysis identifies settings associated with community activity and bullying concentration. Influence analysis further shows that distinct seed configurations can produce comparable diffusion reach under the adopted IC model.

\textbf{Methodological Implications:}
The findings support integrating competency-question-guided ontology design with behavioral and network analysis. The ontology preserves semantic relationships among users, interactions, channel content, discussion subjects, and contextual entities, providing a consistent basis for knowledge graph construction, behavioral representation, community detection, contribution analysis, and influence modeling. ODCBNA thus connects observed corpus records with interpretable network-level findings beyond text-only analysis.

\textbf{Practical Implications:}
The findings suggest that monitoring and mitigation strategies should not rely solely on highly active users or a single centrality measure. Different seed-selection methods identify largely distinct perpetrators while achieving comparable simulated diffusion reach. Combining behavioral similarity, contextual information, network position, and engagement response may therefore help identify potentially influential users within each community.

\textbf{Competency Question Alignment:}
Table~\ref{tab:cq-summary} summarizes how the experimental findings address the competency questions introduced in Section~\ref{subsubsec:cb-onto}, linking the ontology requirements with the analytical outcomes of ODCBNA.

\begin{table*}[ht]
\centering
\caption{Summary of competency questions and corresponding findings.}
\label{tab:cq-summary}
\footnotesize
\setlength{\tabcolsep}{4pt}
\begin{tabular}{c p{5.0cm} p{11.0cm}}
\toprule
\textbf{CQ} & \textbf{Competency Question (Short)} & \textbf{Key Findings} \\
\midrule

CQ1 & Identification of perpetrator users
& Applying the threshold $\tau=3$ identifies $1{,}063$ users with at least three bullying-labeled comments or replies as perpetrators (Section~\ref{subsubsec:perp-subgraph}). \\

CQ2 & Behavioral characteristics of perpetrators
& Community profiles reveal differences in user characteristics, video exposure, engagement response, language use, target identity, political context, temporal activity, and channel exposure (Section~\ref{subsec:community-behavior}). \\

CQ3 & Multi-scale network structure
& The perpetrator similarity network contains $1{,}063$ nodes and $27{,}554$ edges, with an average degree of $51.84$, clustering coefficient of $0.1939$, and average weighted shortest-path distance of $3.4192$ in the largest connected component (Section~\ref{subsec:network-analysis}). \\

CQ4 & Community and sub-community structure
& Louvain detection identifies six communities with a modularity of approximately $0.495$. Hierarchical analysis further reveals $27$ sub-communities, exposing finer structural divisions within the parent communities (Section~\ref{subsec:network-analysis}). \\

CQ5 & Behavior--context association
& Communities differ in behavioral profiles and contextual settings. Attributes associated with greater interaction volume do not necessarily exhibit higher bullying proportions (Section~\ref{subsec:community-interaction}). \\

CQ6 & Contributions to similarity and community separation
& The Video group records the largest discriminative contribution, followed closely by the Impact and User Profile groups. These three groups also dominate the leading individual features, while their relative contributions vary among communities (Section~\ref{subsec:feature-contribution-analysis}). \\

CQ7 & Influential users within communities
& IM Greedy achieves small positive gains in all six communities. Limited seed overlap with the baselines indicates that different methods identify largely distinct potentially influential users (Section~\ref{subsec:im-analysis}). \\

CQ8 & Influence propagation behavior
& Distinct seed sets achieve comparable diffusion reach under the adopted IC settings. Representative trajectories in C5 show similar final activation counts ($182$ versus $181$) but different diffusion depths (four versus five steps) for IM Greedy and Bully-Only (Section~\ref{subsec:im-analysis}). \\

\bottomrule
\end{tabular}
\end{table*}

\subsection{Limitations}
\label{subsec:limitations}

Several limitations should be considered when interpreting the findings.

\textbf{Behavioral Representation:}
The analysis uses a manually designed $64$-dimensional behavioral--contextual feature vector and cosine similarity, which may not capture latent or nonlinear behavioral relationships. The contribution results depend on the selected attributes, preprocessing decisions, and feature-group definitions. Learned or hybrid representations may capture additional patterns while preserving interpretability.

\textbf{Analytical Granularity:}
Behavioral, contextual, and contribution analyses are conducted mainly at the community level. Although $27$ sub-communities are identified, these analyses are not systematically extended to that finer scale. Local behavioral variations within the parent communities may therefore remain underrepresented.

\textbf{Network and Model Parameters:}
The perpetrator set, similarity network, community structure, and diffusion outcomes depend on methodological choices such as the perpetrator threshold $\tau$, neighborhood size $k_{\mathrm{NN}}$, Louvain resolution, seed budget $k_s$, and IC propagation parameters. The findings therefore reflect the adopted experimental settings rather than parameter-optimal configurations. Further sensitivity analysis is needed to assess the stability of these results under alternative parameter values.

\textbf{Temporal and Diffusion Assumptions:}
The similarity network is treated as static, and the IC model uses fixed edge-level activation probabilities. However, user behavior and interaction patterns may evolve over time. Therefore, the diffusion results represent simulated outcomes under the adopted network and parameter settings rather than direct estimates of real-world propagation.

\textbf{Dataset Scope and Generalizability:}
The framework is instantiated using video-based Bengali political-news interactions collected from selected YouTube channels during a specific period. The findings may depend on the language, topic, collection window, and source-selection criteria of BenCoDiC. Evaluation using other YouTube content types, languages, topics, and time periods is needed to assess broader generalizability.

These limitations motivate future work on learned or hybrid representations, sub-community-level analysis, parameter sensitivity, temporal network modeling, and cross-domain evaluation.

\section{Conclusion}
\label{sec:conclusion}

This paper presented an Ontology-Driven Cyberbullying Behavior and Network Analysis (ODCBNA) framework for perpetrator-centric cyberbullying analysis on YouTube. Guided by analytical requirements and competency questions, ODCBNA integrates application ontology design, knowledge graph construction, heterogeneous graph transformation, behavioral similarity modeling, community analysis, and influence propagation within a unified pipeline.

Using BenCoDiC as a use case, the framework identified $1{,}063$ perpetrators and constructed a weighted similarity network with $27{,}554$ edges. The network revealed six communities with a modularity of approximately $0.495$ and $27$ sub-communities, demonstrating organization at multiple structural levels. The Video group recorded the largest discriminative contribution, followed closely by the Impact and User Profile groups. These groups also dominated the leading individual features, while the Language, Target, Party, Time, and Channel groups remained useful for contextual interpretation.

The influence experiments produced closely comparable spreads among the four seed-selection methods. IM Greedy achieved small positive gains in all six communities, despite limited seed overlap with the baselines. These findings suggest that distinct seed configurations can achieve comparable diffusion reach under the adopted IC settings.

Overall, ODCBNA establishes a traceable connection between heterogeneous corpus records, semantic relationships, behavioral representations, network structure, and influence propagation. The findings suggest that cyberbullying analysis and intervention should consider behavioral, contextual, and structural evidence rather than relying solely on textual content, user activity, or individual centrality measures.

Future work will examine learned and hybrid representations, parameter sensitivity, temporal network modeling, and sub-community-level analysis. It will also incorporate semantically linked external data descriptions to extend the analytical capabilities of ontologies and knowledge graphs. Further evaluation using other languages, topics, time periods, and social platforms will assess the framework's broader generalizability.

\printcredits

\section*{Declaration of competing interest}

The authors declare that they have no known competing financial interests or personal relationships that could have appeared to influence the work reported in this paper.

\bibliographystyle{cas-model2-names}

\bibliography{references}





\end{document}